\documentclass[lettersize,journal]{IEEEtran}
\usepackage{amsmath,amsfonts}
\usepackage{algorithmic}
\usepackage{array}
\usepackage[caption=false,font=footnotesize,labelfont=rm,textfont=sf]{subfig}
\usepackage{textcomp}
\usepackage{stfloats}
\usepackage{url}
\usepackage{verbatim}
\usepackage{graphicx}
\usepackage{cite}
\usepackage{booktabs}
\usepackage{threeparttable}
\usepackage{color,xcolor}
\usepackage{amssymb}
\usepackage{multirow}
\usepackage[ruled, vlined]{algorithm2e}

\newcommand*{\red}{\textcolor{black}}

\begin{document}

\title{ODMixer: Fine-grained Spatial-temporal MLP for Metro Origin-Destination Prediction}

\author{Yang Liu,~\IEEEmembership{Member,~IEEE},  Binglin Chen, Yongsen Zheng, Lechao Cheng, Guanbin Li, Liang Lin,~\IEEEmembership{Fellow,~IEEE}
\thanks{This work is supported by the National Key R\&D Program of China under Grant No.2021ZD0111601, in part by the National Natural Science Foundation of China under Grant No.62002395, in part by the Guangdong Basic and Applied Basic Research Foundation under Grant No.2023A1515011530, and in part by the Guangzhou Science and Technology Planning Project under Grant No. 2023A04J2030. (\emph{Corresponding author: Liang Lin.})} 
\thanks{Yang Liu, Binglin Chen, Yongsen Zheng, Guanbin Li and Liang Lin are with the School of Computer Science and Engineering, Sun Yat-sen University, Guangzhou, China. (E-mails: liuy856@mail.sysu.edu.cn; chenblin9@mail2.sysu.edu.cn; z.yongsensmile@gmail.com; liguanbin@mail.sysu.edu.cn; linliang@ieee.org)}
\thanks{Lechao Cheng is with the School of Computer Science and Information
Engineering, Hefei University of Technology, Hefei, 230601, China. (e-mail: chenglc@hfut.edu.cn)}}

\markboth{IEEE Transactions on Knowledge and Data Engineering}%
{Shell \MakeLowercase{\textit{et al.}}: A Sample Article Using IEEEtran.cls for IEEE Journals}


\maketitle

\begin{abstract}
\red{
Metro Origin-Destination (OD) prediction is a crucial yet challenging spatial-temporal prediction task in urban computing, which aims to accurately forecast cross-station ridership for optimizing metro scheduling and enhancing overall transport efficiency. Analyzing fine-grained and comprehensive relations among stations effectively is imperative for metro OD prediction. However, existing metro OD models either mix information from multiple OD pairs from the station's perspective or exclusively focus on a subset of OD pairs. These approaches may overlook fine-grained relations among OD pairs, leading to difficulties in predicting potential anomalous conditions. To address these challenges, we learn traffic evolution from the perspective of all OD pairs and propose a fine-grained spatial-temporal MLP architecture for metro OD prediction, namely ODMixer. Specifically, our ODMixer has double-branch structure and involves the Channel Mixer, the Multi-view Mixer, and the Bidirectional Trend Learner. The Channel Mixer aims to capture short-term temporal relations among OD pairs, the Multi-view Mixer concentrates on capturing spatial relations from both origin and destination perspectives. To model long-term temporal relations, we introduce the Bidirectional Trend Learner. Extensive experiments on two large-scale metro OD prediction datasets HZMOD and SHMO demonstrate the advantages of our ODMixer. Our code is available at https://github.com/KLatitude/ODMixer.
}

\end{abstract}


\section{Introduction}

\begin{figure}
\centering
    \subfloat[Different Views for Embedding OD Matrix]{
        \includegraphics[width=80mm]{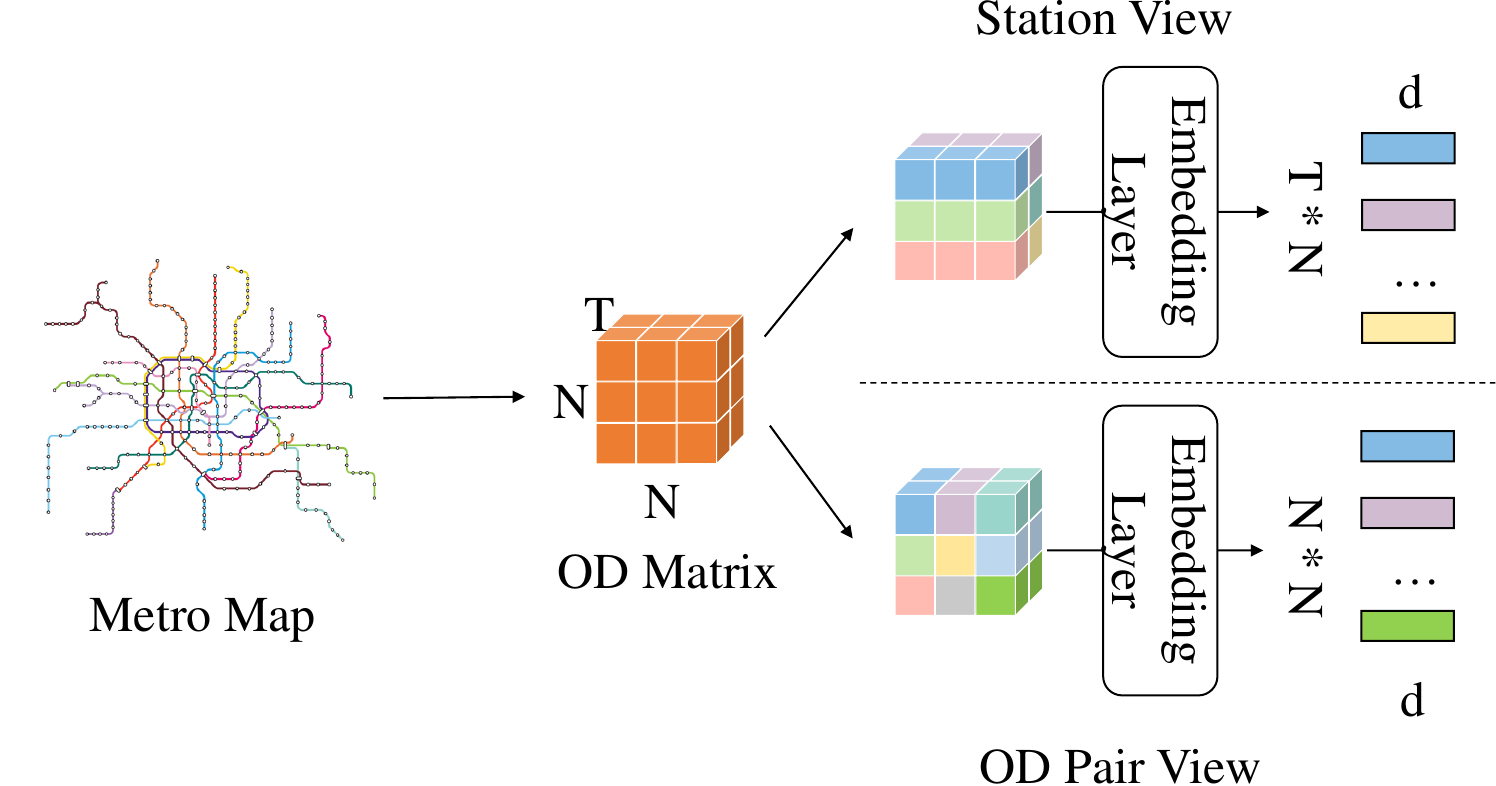}
    }\
    \\
    \subfloat[Different Models for Dealing $N^2$ Tokens]{
        \includegraphics[width=80mm]{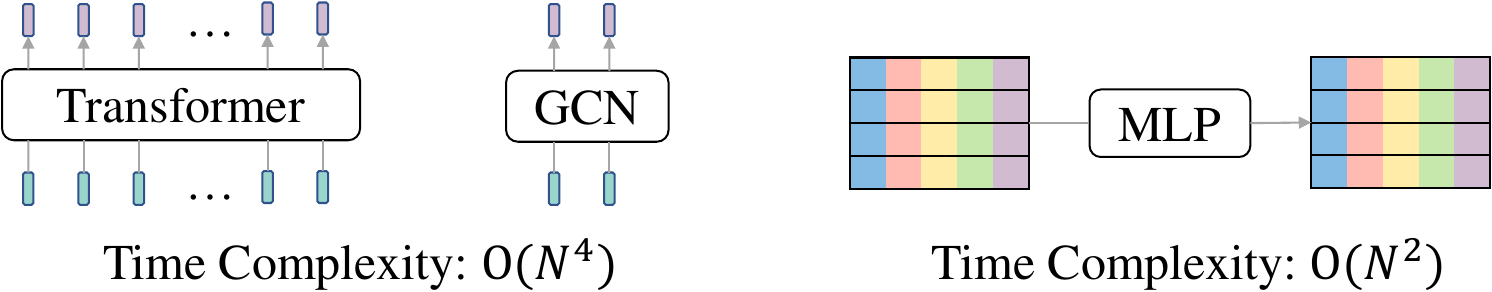}
    }
\caption{Comparison of different views and models for metro OD prediction. (a) illustrates the difference in encoding the OD Matrix from the station view and the OD pair view. N is the number of stations, T is the number of time intervals. (b) shows various models for processing the $N^{2}$ tokens.}
\label{fig:introduction}
\end{figure}


With the rapid development of urbanization, city populations are surging, leading to a gradual increase in the number of vehicles \cite{jin2023spatio,wang2023urban,ijcai2023p672,li2024survey}. To enhance daily commuting convenience, metro is increasingly considered as the preferred transportation mode. Concurrently, cities are proactively expanding their metro networks to augment transport capacity and alleviate traffic pressures, thereby enhancing the quality of life for urban residents. In some metropolises, such as Beijing and Shanghai, nearly 10 million metro travel transactions occur daily. This immense ridership poses significant challenges for metro operations. Consequently, accurately forecasting future ridership is crucial for effective metro scheduling and route planning. However, to ensure the operational efficiency of the metro system and the seamless movement of passengers, real-time monitoring of metro passenger flow and predicting future trends have become imperative to address various emergencies. Consequently, metro OD prediction, directed at estimating the flow among all stations in the forthcoming period, emerges as a critical task in metro scheduling and flow management.


\red{
Metro OD prediction poses unique challenges compared to general OD prediction \cite{zheng2022metro,liu2022online,gong2020online}. First, the time gap between passenger entry and exit often causes these events to occur in different time intervals. As a result, the OD matrix for the current time interval cannot fully represent passengers still within the system, leading to incompleteness. Second, the high number of metro stations results in a high-dimensional OD matrix, complicating model design. Additionally, uneven passenger flow distribution creates significant disparities between OD pairs, contributing to matrix sparsity. These factors make metro OD matrix contain highly complex spatiotemporal relations, making metro OD prediction exceptionally challenging.}

\begin{figure}
    \centering
\subfloat[\scriptsize{Flow difference between OD pairs}]
        {
\includegraphics[width=40mm]{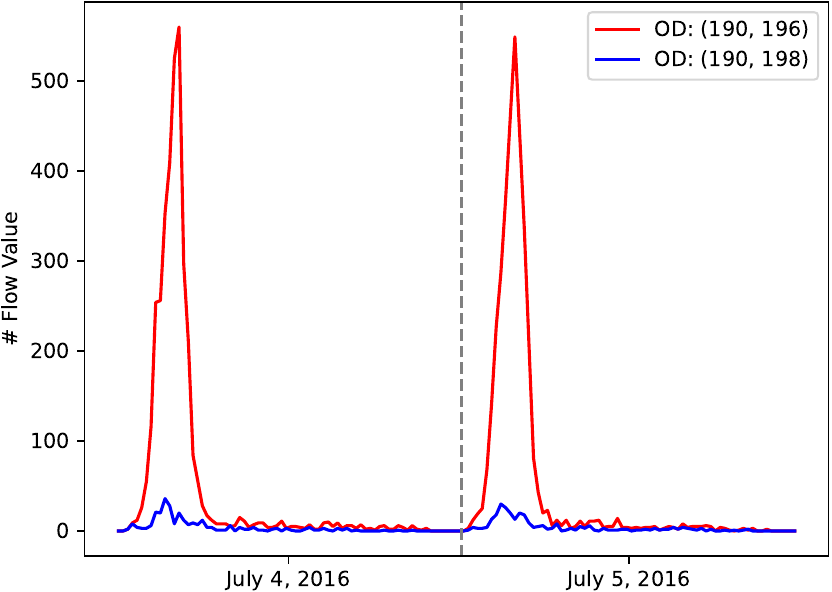}
    }
    \vspace{-0.11em}
\subfloat[\scriptsize{Trend difference between OD pairs}]{
        \includegraphics[width=40mm]{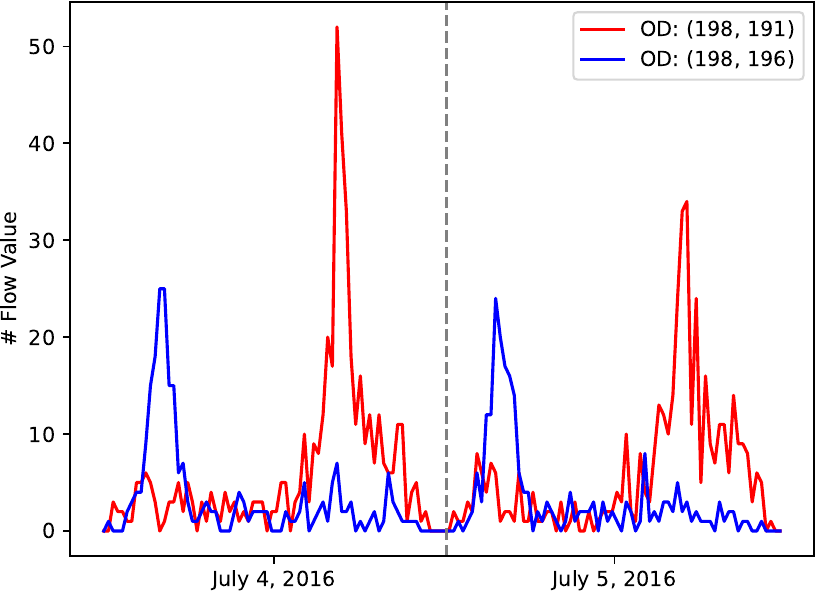}
    }
    \caption{Comparison of flow changes between OD pairs with the same origin. The x-axis represents July 4 and 5, 2019, the y-axis represents the flow value between OD pairs. The OD: (190,196) in the legend means the origin is station 190 and the destination is station 196. Figure (a) shows that although the OD pairs with the same origin have similar change trends, the flow values are very different. Figure (b) illustrates that the flow rate trends of two OD pairs with the same origin exhibit significant differences over time.}
    \vspace{-15pt}
\label{fig:od_flow}
\end{figure}

\red{For metro OD prediction, recent years have seen the rise of deep learning models leveraging powerful nonlinear modeling capabilities \cite{liu2022online,gong2020online,xu2023adaptive,shen2024short}. These models employ architectures such as RNNs, TCNs, and Transformers to capture temporal relations, while GCNs and Transformers are commonly utilized to learn spatial dependencies. However, most of these approaches adopt a station-based perspective, as illustrated in Fig. \ref{fig:introduction}(a). This method encodes data for each station during each time interval, causing the flow data of different OD pairs with the same origin to be mixed at the input layer. As a result, the extracted features combine information from multiple OD pairs, making it challenging for the model to distinguish between them when learning spatiotemporal relations. This limitation makes the station-based perspective a coarse-grained modeling method.
As depicted in Fig. \ref{fig:od_flow}(a), while the overall flow trends of two OD pairs with the same origin may appear similar over time, significant differences in their flow values, particularly during peak hours, are evident. Moreover, in some cases, as shown in Fig. \ref{fig:od_flow}(b), the flow trends of OD pairs with the same origin can be entirely opposite. Mixing these OD pairs at the input layer overlooks such differences, hindering the model’s ability to effectively learn spatiotemporal relations.
To address this, we propose an OD pair-based modeling perspective, as shown in the Fig. \ref{fig:introduction}(a). This approach models each OD pair independently, allowing the model to differentiate between OD pairs and accurately capture their complex spatiotemporal relations. Compared to the station-based perspective, the OD pair-based modeling approach provides a fine-grained representation of the OD matrix, enabling the model to learn more fine-grained and accurate patterns in the data. Additionally, from Fig. \ref{fig:od_flow}(a) and (b), we can see that for each OD pair, the OD traffic at adjacent day exhibits a very similar trend. However, existing methods \cite{liu2022online} do not consider this similarity and fail to incorporate historical traffic information, leading to sub-optimal model performance.
}

Actually, there exist some models that predict the metro OD matrix from the OD pair view \cite{zhang2023deep,yang2022spatiotemporal}, but they only focus on a subset of OD pairs in the metro. This constrained approach neglects the richness of information among all OD pairs, resulting in the model's inability to learn fine-grained relations across the entirety of OD pairs. Consequently, it hampers the model's capability to address potential emergencies and comprehend the dynamics of traffic changes, making it difficult to provide effective support for metro management. To achieve more comprehensive metro OD prediction, it is imperative to consider information from all OD pairs. However, considering the number of OD pairs as $N^2$, the time complexity needed to calculate the relations among $N^2$ OD pairs using Transformer or to construct a virtual graph among OD pairs and subsequently employ Graph Convolution Network (GCN) for information aggregation is $O(N^4)$, as shown in Fig. \ref{fig:introduction}(b). Therefore, balancing the model's parameters and computation time while considering all OD pairs remains challenging.


\red{
Inspired by the efficiency of the MLP architecture \cite{tolstikhin2021mlp,lin2024mlp}, we propose ODMixer, a simple yet effective fine-grained spatiotemporal MLP architecture. Unlike existing metro OD prediction models that adopt coarse-grained station-based modeling, ODMixer employs a fine-grained OD pair-based modeling approach to comprehensively capture the complex spatial-temporal relations in metro OD prediction. To address the issue of incomplete input data in metro OD prediction, we introduce the OD Matrix Process module, which leverages periodicity to fill missing values in the OD matrix. Once the OD matrix is complete, we independently model each OD pair from an OD pair-based perspective. This approach not only considers each OD pair separately but also mitigates the challenges posed by uneven distribution in the OD matrix. To capture the intricate spatiotemporal relationships within the OD matrix, we design specialized modules. For temporal relations, we account for both short-term and long-term dependencies. The Channel Mixer operates on the most recent features of each OD pair to capture short-term temporal relations. To model long-term trends, such as periodicity inherent in metro traffic, we employ a dual-branch structure coupled with a Bidirectional Trend Learner. Regarding spatial relations, we treat each OD pair as a node. Given the computational challenges posed by models like Transformers and GCNs in processing all OD pairs, we adopt a decomposition strategy. Specifically, we decompose spatial relations based on the origin and destination perspectives and design the Multi-view Mixer module. This module separately calculates spatial relations from the origin and destination perspectives, utilizing a Mixer structure for each. This design enhances the model’s efficiency and its ability to learn spatial relationships across all OD pairs.
}
Our contributions are summarized as follows:
\begin{itemize}
    \item We propose a fine-grained spatial-temporal MLP architecture named ODMixer from the perspective of the OD pairs, to comprehensively capture OD relations and achieve accurate and efficient metro OD prediction.
    \item To effectively learn the spatial and temporal dependencies between metro flows, we propose two specific modules, the Channel Mixer and the Multi-view Mixer.
    \item To empower the model with the capability to perceive long-term flow changes, the Bidirectional Trend Learner (BTL) is introduced in the ODMixer.
    \item Extensive experiments on two large-scale metro OD datasets demonstrate our promising OD prediction accuracy. Our ODMixer outperforms the state-of-the-art models in wMAPE for HZMOD and SHMOD datasets by a considerable margin of 5\% and 7\%, respectively.
\end{itemize}

\section{Related Work}
\subsection{Traffic Flow Prediction}
Traffic flow prediction is a very important task in smart city systems. Its main task is to use historical traffic information to accurately predict future traffic flow. There are many methods to solve this problem. Initially, researchers used traditional statistical methods to analyze traffic changes, such as ARIMA\cite{williams1998urban}. Later, some machine learning-based methods began to emerge, such as SVM\cite{sun2015novel} and decision trees\cite{elleuch2017intelligent}.

With the rapid development of deep learning, neural networks have been used to solve traffic prediction problems. These representative works include many structural types, mainly including CNN \cite{zeng2021modeling,qu2022forecasting}, RNN \cite{zhaowei2020short,ma2021short}, Attention \cite{guo2019attention,hu2021attention}, etc. DeepSD\cite{wang2017deepsd} proposed an end-to-end deep learning framework and integrates external environment information to realize the automatic discovery of complex supply-demand patterns from the car-hailing service data. Based on three temporal relations, MDL\cite{zhang2019flow} used two three-stream fully convolutional networks (3S-FCNs) to achieve flow prediction for points and edges. DeepCrowd\cite{jiang2021deepcrowd} divided the area into multiple fine-grained grids, and combines ConvLSTM and attention mechanisms to analyze and fuse flow information at multiple times to achieve Large-Scale Citywide Crowd Density and Flow Prediction. Later, the GCN was popular to process the relation between nodes because of the natural network structure in the traffic flow prediction problem \cite{wang2020traffic,zhang2021traffic,ou2022stp}. STDGRL\cite{xie2023spatio} designed a spatio-temporal dynamic graph relation learning model. STWave+\cite{fang2023stwave+} utilized decoupling technology to partition one-hour traffic data into stable trends and fluctuating events. This approach employs a dual-channel spatiotemporal network to separately model these components, leveraging self-supervised learning and contrastive loss to propagate long-term trend information to hourly trends. 

However, these methods focus solely on the total flow at each node, they fail to analyze the flow relations between nodes. This limitation restricts the information that can be obtained, making it less suitable for practical applications such as traffic analysis and control. To address this issue, we concentrate on the relations between nodes. Compared to focusing only on the total flow at each node, this OD relation provides a more fine-grained analysis. Additionally, the total flow at a node can be derived from the OD flow, highlighting the greater value of OD flow data.

\subsection{Origin-Destination Prediction}

OD prediction \cite{wang2024traffic,huang2023odformer,he2023short,jiang2021countrywide} involves the accurate estimation of traffic flow between two regions over a given period of time, presenting a formidable challenge due to its intricate and dynamic nature. As a pivotal aspect of urban traffic management, OD prediction has garnered extensive attention.


As a special type of OD prediction, the metro OD prediction \cite{gong2020online,zhang2021short,liu2022online,ye2023completion,xu2023adaptive,zhang2023deep,zhu2023two,zheng2022metro} presents unique challenges compared to the general OD prediction, primarily because the precise destinations of passengers are not accurately known until they reach their destinations. This leads to incomplete data at the current moment, making it challenging to effectively depict the real-time traffic distribution. To address this issue and obtain more comprehensive data on traffic distribution, several studies focused on predicting unfinished orders to supplement the incomplete OD matrix. For instance, in HIAM\cite{liu2022online}, long and short historical distributions were utilized to estimate the distribution of unfinished orders at the current time. This study integrated the completed and unfinished distributions in the feature space. Gong et al. \cite{gong2020online} created an indication matrix, with values assigned based on the assumption that the travel time of each OD pair at each timestamp follows a normal distribution. However, this approach overlooked the unfinished OD pair data. MVPF \cite{zheng2022metro} proposed a multi-view passenger flow evolution trend based OD matrix prediction method that combines individual station and cross-station learning, uses GRU and EGAT models to learn the spatial-temporal-dependent representation of stations, and defines a transfer matrix to capture passenger mobility patterns.

Different from previous approaches that neglect the richness of information among all OD pairs, we consider all OD pairs from the OD pair view and propose a novel model base on the spatial-temporal MLP architecture to achieve more fine-grained and comprehensive relation modeling.

\subsection{MLP-based Model}
With the popularity of the Transformer, its exceptional performance has led to its widespread adoption across various fields. However, several MLP-based models have attracted attention for addressing diverse problems recently. In computer vision, MLP-Mixer \cite{tolstikhin2021mlp} addressed vision challenges through an entirely MLP-based implementation. Another notable model gMLP \cite{liu2021pay} outperformed some Transformer-based models by leveraging channel projection, spatial projection, and gating mechanisms.  ResMLP\cite{touvron2022resmlp} introduced a residual structure atop MLP, incorporating training methods like self-supervision for efficient training.  sMLPNet\cite{tang2022sparse} focused on reducing computational costs while maintaining performance. For sequential recommendation problems, MLP4Rec\cite{li2022mlp4rec} employed a tri-directional mixing MLP model, while MLPST\cite{zhang2023mlpst} introduced SpatialMixer and TemporalMixer to address spatio-temporal dependencies and temporal variations in traffic prediction. For time series forecasting task, Ekambaram et al.\cite{ekambaram2023tsmixer} and Chen et al.\cite{chen2023tsmixer} proposed MLP-based models dedicated to efficient time series forecasting.

\begin{figure*}[!t]
    \centering
\includegraphics[width=160mm]{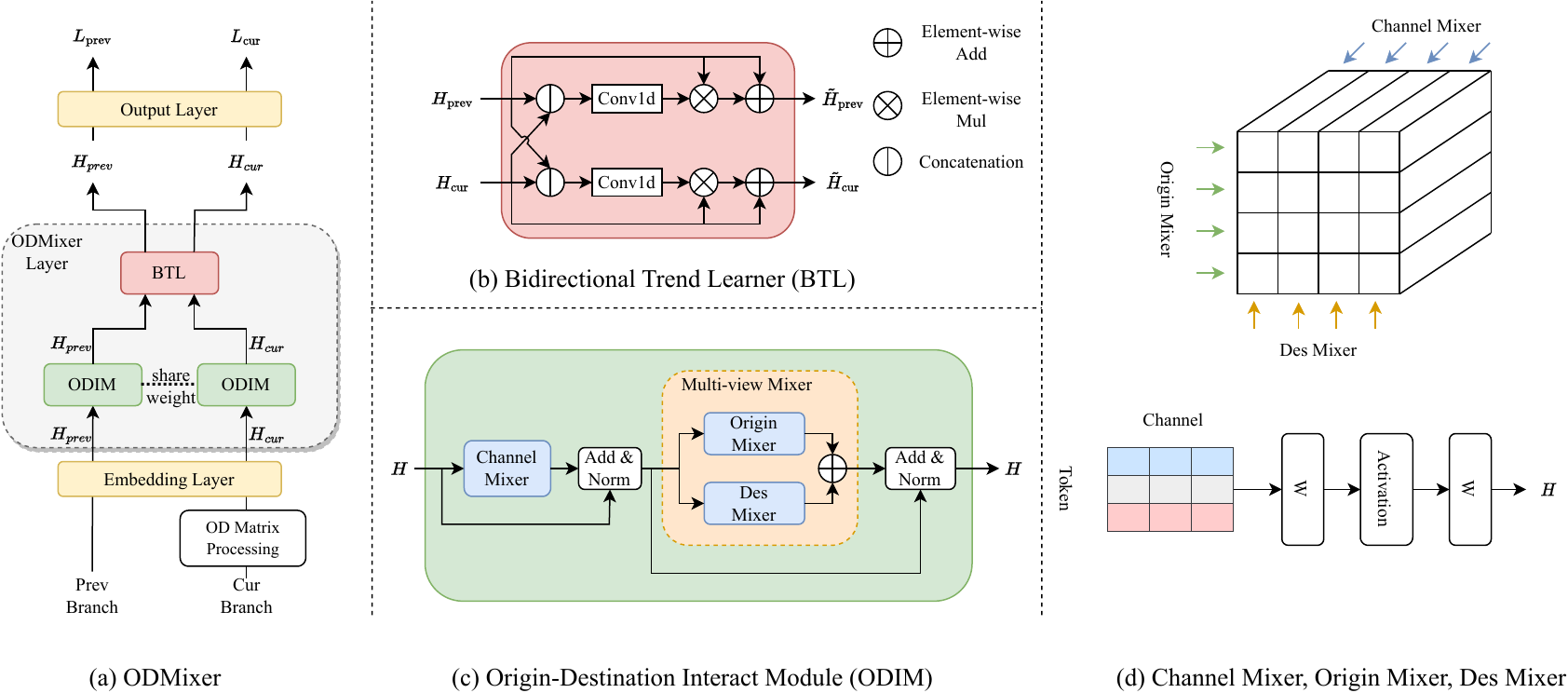}
\vspace{-10pt}
    \caption{The overall framework of the ODMixer and its essential modules. (a) illustrates the architecture of double-branch ODMixer, both branches learn features using ODIM. Subsequently, the two branches interact with each other using the BTL, contributing to the final output. (b) depicts the BTL module, which aims to jointly model the features from both branches, thereby enhancing the information exchange. (c) represents the ODIM module, which models the temporal attributes of the OD pair features using the Channel Mixer, while the comprehensive relations among OD pairs are learned using the Multi-view Mixer. (d) shows the Channel Mixer, Origin Mixer, Des Mixer.}
    \vspace{-10pt}
    \label{fig:overview}
\end{figure*}

Unlike existing MLP-Mixer models that neglect fine-grained spatial-temporal relations, we focus on metro OD prediction task that requires fine-grained and comprehensive OD relations. In this paper, we propose a fine-grained spatial-temporal MLP model ODMixer, which excels in capturing spatial relations of metro flows, along with short-term and long-term temporal dependencies.

\section{Preliminaries}
In this section, a concise introduction is provided for the notation and problem definitions. Table \ref{table:notation} presents a comprehensive list of commonly used symbols for reference.

\textbf{Definition 1.} \red{Incomplete OD Matrix $\mathrm{IOD}_{t}^d \in \mathbb{R}^{N\times N}$: $\mathrm{IOD}_{t}^d(i, j)$ is the number of passengers entering from station $i$ and exiting from station $j$ at the $t^{th}$ time interval of the $d^{th}$ day.}

\textbf{Definition 2.} \red{Unfinished OD Matrix $\mathrm{UOD}_{t}^d \in \mathbb{R}^{N\times N}$: $\mathrm{UOD}_{t}^d(i, j)$ signifies the number of passengers entering station $i$ at the $t^{th}$ time interval and exiting from station $j$ after the $t^{th}$ time interval of the $d^{th}$ day.}

\textbf{Definition 3.} \red{Complete OD Matrix $\mathrm{OD}_{t}^d \in \mathbb{R}^{N\times N}$: $\mathrm{OD}_{t}^d(i, j)$ denotes the number of passengers at the $t^{th}$ time interval entering station $i$ and subsequently exiting from station $j$ at and after the $t^{th}$ time interval of the $d^{th}$ day. $\mathrm{OD}_{t}^d = \mathrm{IOD}_{t}^d + \mathrm{UOD}_{t}^d$.}

\textbf{Definition 4.} \red{Unfinished Order Vector $\mathrm{unf}_{t}^d \in \mathbb{R}^{N}$: $\mathrm{unf}_{t}^d(i)$ represents the number of passengers at the $t^{th}$ time interval entering station $i$ but not exiting from any station at the $t^{th}$ time interval of the $d^{th}$ day. $\mathrm{unf}_{t}^d(i) = \sum_{j=1}^{N}\mathrm{UOD}_{t}^d(i, j)$.}

\textbf{Problem 1.} Considering the historical complete OD matrix (OD), alongside the incomplete OD matrix (IOD) and the unfinished orders (unf) from the most recent time period, we determine a function $f$ to predict the complete OD flow at the subsequent time step.
\begin{equation}
\mathrm{OD}_{t+1}^{d} = f(\{\mathrm{OD}^{d-1}, \mathrm{IOD}^{d}, \mathrm{unf}^{d}\}_{t-T+i})
\end{equation}
where $i=1, 2, \cdots T$, $OD_{t+1}^d$ represents the OD matrix at the $(t+1)^{th}$ time interval of the $d^{th}$ day.

\section{Methodology}

This section gives a detailed description of our ODMixer.

\begin{table}[!t]\scriptsize
\caption{Notation Table.}
\vspace{-5pt}
\centering
\begin{tabular}{c|c}
\toprule
Notations & Description \\
\midrule
$t$ & the current time interval \\
$d$ & the current day \\
$T$ & the number of time intervals of input \\
$N$ & the number of metro stations \\
\red{$\mathrm{unf}_{t-T+i}^d \in \mathbb{R}^{N}$} & \red{the unfinished order vector ($i = 1, \cdots, n$)} \\
\red{$\mathrm{OD}_{t-T+i}^d \in \mathbb{R}^{N \times N}$}  & \red{the complete OD matrix ($i = 1, \cdots, n$)} \\
\red{$\mathrm{UOD}_{t-T+i}^d \in \mathbb{R}^{N \times N}$} & \red{the unfinished OD matrix ($i = 1, \cdots, n$)} \\
\red{$\mathrm{IOD}_{t-T+i}^d \in \mathbb{R}^{N \times N}$} & \red{the incomplete OD matrix ($i = 1, \cdots, n$)} \\
\bottomrule
\end{tabular}
\vspace{-10pt}
\label{table:notation}
\end{table}

\subsection{Architecture Overview}
Our proposed ODMixer comprises two branches that operate similarly and share parameters. Initially, in the Embedding Layer, we embed the input based on the OD pair view. Subsequently, we employ ODIM to learn the relations between features. Specifically, we utilize Channel Mixer to learn the short-term temporal relations of features, followed by employing Multi-view Mixer to comprehensively learn the relations among OD pairs from both Origin and Destination perspectives. Subsequently, we use BTL to interact the features of the two branches, enhancing the long-term temporal attributes of the features. This augmentation ensures that the features contain more information, ultimately contributing to increase prediction accuracy. Finally, the Output Layer is employed to output the features.

\subsection{Dual-branch Structure}

As shown in Fig. \ref{fig:od_flow}, the OD flow on adjacent days exhibits certain similarities. Therefore, we integrated this similarity into our model by designing a dual-branch structure. The Prev Branch learns historical information, while the Cur Branch learns information at the current time. Since we focus on the similarity between adjacent days, the input to the Prev Branch is the OD matrix from the same time on the previous day. Since yesterday's order information has been completed, its input is a complete OD matrix and no operation is required. For the Cur branch, the OD matrix is incomplete as some orders have not yet been completed.

If we directly use the incomplete OD matrix, discrepancies between the inputs of the Prev Branch and Cur Branch arise, hindering the model's ability to effectively perceive complete information for the current time. Therefore, we introduce the OD Matrix Processing module in the Cur branch.

Specifically, we use an Embedding layer to encode the OD matrix based on the OD pair perspective for the inputs of both branches. Subsequently, the ODIM module learns the relations between OD pairs. We then introduce the BTL module to learn the relations between the two branches, supplementing the Cur Branch with historical information. This design captures the similarities between adjacent days and completes the incomplete information for the current time, improving the model's prediction accuracy and reliability.

This dual-branch structure not only manages the complex relation between OD traffic and time but also enhances the prediction accuracy of current OD traffic by leveraging historical information. Consequently, our model can better handle the large and complex OD matrix in the subway system, utilizing historical data for accurate prediction, thereby demonstrating significant potential and practical value in real-world applications.

\subsection{OD Matrix Processing}
In this subsection, we introduce the approach for supplementing the OD matrix and estimating the current UOD distribution by leveraging the historical UOD distribution.

\subsubsection{Unfinished Orders Processing}
As the destination information is only available until passengers have exited the station, the OD matrix for the current time period is incomplete. To address this issue and enhance the completeness of the input OD matrix for the model, preprocessing steps are implemented before integrating the matrix into the model.

Acknowledging the temporal similarity characteristic of metro traffic, where the current OD traffic distribution closely resembles that of the previous time, we estimate the current UOD distribution by referencing historical UOD distributions. Our approach specifically incorporates two temporal scales: long-term and short-term UOD distributions. The calculation is outlined as follows:
\begin{equation}\small
\begin{aligned}
\red{\mathrm{US}^{d}_{t-T+i}(j, k)} &\red{=} \red{\mathrm{unf}^{d}_{t-T+i}(j) \ast \frac{\mathrm{UOD}^{d-1}_{t-T+i}(j, k)}{\sum_{k=1}^{N}{\mathrm{UOD}^{d-1}_{t-T+i}(j, k)}}} \\
\red{\mathrm{UL}^{d}_{t-T+i}(j, k)} &\red{=} \red{\mathrm{unf}^{d}_{t-T+i}(j) \ast \frac{\mathrm{UOD}^{d-7}_{t-T+i}(j, k)}{\sum_{k=1}^{N}{\mathrm{UOD}^{d-7}_{t-T+i}(j, k)}}} \\
\red{\widetilde{\mathrm{UOD}}^{d}_{t-T+i}(j, k)} &\red{=} \red{\frac{1}{2}(\mathrm{US}^{d}_{t-T+i}(j, k) + \mathrm{UL}^{d}_{t-T+i}(j, k))} \\
\end{aligned}
\end{equation}
\red{Here, $i=1,\cdots,T$, $\widetilde{\mathrm{UOD}}^{d}_{t-T+i}$ represents an estimation of unfinished OD matrix at the ${(t-T+i)}^{th}$ time interval of the $d^{th}$ day, the symbol $\ast$ denotes the multiplication of the corresponding elements. $\mathrm{UOD}^{d-1}_{t-T+i}$ represents the short-term UOD distribution, corresponding to the UOD distribution at the same time yesterday, and $\mathrm{UOD}^{d-7}_{t-T+i}$ represents the long-term UOD distribution, aligned with the UOD distribution observed at the corresponding time on the same day of the previous week. This computation aims to generate a more comprehensive and accurate estimate of the UOD distribution at the current time by combining both short-term and long-term UOD distributions.}

Subsequently, we derive the OD matrix for the current time, which serves as input to the Embedding Layer:
\begin{equation}
\red{\widetilde{\mathrm{OD}}^{d}_{t-T+i}(j, k) = \mathrm{IOD}^{d}_{t-T+i}(j, k) + \widetilde{\mathrm{UOD}}^{d}_{t-T+i}(j, k)}
\end{equation}
\red{where $i=1,\cdots,T$, $\widetilde{\mathrm{OD}}^{d}_{t-T+i}$ is the estimation of complete OD matrix at the current time, $\mathrm{IOD}^{d}_{t-T+i}$ represents the matrix of actual completed orders, and $\widetilde{\mathrm{UOD}}^{d}_{t-T+i}$ denotes the matrix of unfinished orders computed from the historical short-term and long-term UOD distributions. This approach ensures the model receives comprehensive and accurate input information by integrating data from both completed and ongoing orders, thereby enhancing the accuracy of passenger destination predictions.}

\subsection{Embedding Layer}
Unlike the previous models for handling OD matrix, we consider it from the OD pair view. We encode each OD pair sequence using the Embedding Layer, which ultimately yields fine-grained features of the OD matrix:
\begin{equation}
\red{\boldsymbol{H}(i, j) = \boldsymbol{W^E}\mathrm{OD}_{t-T+1:t}(i, j)}
\end{equation}
\red{where $i = 1,\cdots,N, j=1,\cdots,N$, $\boldsymbol{W^E} \in \mathbb{R}^{d \times T}$ is learnable weights of Embedding layer, $\boldsymbol{H} \in \mathbb{R}^{N \times N \times d}$ represents the obtained feature representation after the Embedding Layer. This approach differs from the previous methods that treat the OD matrix as a whole. Instead, we emphasize modeling each OD pair independently, to effectively capture the characteristics and dynamics of each OD pair. After performing embedding layer on both Prev and Cur branches, we obtain the feature $\boldsymbol{H_{\mathrm{prev}}}, \boldsymbol{H_{\mathrm{cur}}} \in \mathbb{R}^{N \times N \times d}$}.

By modeling OD sequences, we can distinguish the difference among OD pairs at a fine-grained manner, which enhances the sensitivity and accuracy of the model in predicting OD matrix. Moreover, this approach is more realistic, considering significant differences among various OD pairs. Modeling these OD pairs individually better reflects the inherent variability within the data.

\subsection{Origin-Destination Interact Module (ODIM)}
For the metro OD prediction task, the spatio-temporal relations plays a crucial role. First, passenger flow information in adjacent time periods is correlated, indicating that current passenger flow can be influenced by flows in preceding and succeeding periods. Second, each metro station has unique attributes, and stations with similar attributes typically exhibit similar passenger flow trends. Therefore, by learning and capturing the spatial relations of OD pairs, the model's prediction performance can be significantly enhanced, leading to more accurate predictions of future passenger flow distribution. To address this, we designe the ODIM module to effectively capture the spatio-temporal relations of OD pairs. ODIM conducts spatio-temporal interaction of features encoded by the Embedding Layer. It comprises two modules: Channel Mixer, which considers the short-term temporal relations of features within OD pairs, and Multi-view Mixer, which models the relations among OD pairs.

\subsubsection{Channel Mixer}
In the domain of metro traffic, temporal attributes play a pivotal role, and there exist discernible correlations between traffic patterns at different time. Consequently, the model must adeptly capture the temporal characteristics of flow to ensure accurate OD predictions. Leveraging the Embedding Layer, we encode each OD sequence as features. To effectively capturing the time-varying relations within the flow, it is imperative to compute the inter-dependencies among these features. Thus, we introduce the Channel Mixer structure, facilitating the fusion of attributes across different time to discern and learn the temporal patterns inherent in the flow. This yields a feature representation enriched with temporal attributes.

The Channel Mixer structure can be expressed as follows:
\begin{equation}
\boldsymbol{H^{C}_{i}} = LayerNorm(\boldsymbol{H_i} + \boldsymbol{W^{C_2}} \sigma(\boldsymbol{W^{C_1}}\boldsymbol{H_i}))
\end{equation}
where, $i = 1, \cdots, N^2$, $\sigma$ is the activation function, $\boldsymbol{W^{C_1}} \in \mathbb{R}^{d_{C} \times d}$ is learnable weights of the first fully connected layer in the Channel Mixer, $\boldsymbol{W^{C_2}} \in \mathbb{R}^{d \times d_{C}}$ is learnable weights of the second fully connected layer, $\boldsymbol{H^C} \in \mathbb{R}^{N \times N \times d}$ represents the feature after temporal attributes have enhanced by the Channel Mixer. We employ layer normlization (LayerNorm) \cite{ba2016layer} and residual connection \cite{he2016deep} in the Channel Mixer. The Channel Mixer structure is specifically designed to merge attributes from diverse time intervals within the feature representation, enhancing the model's ability to capture temporal traffic patterns. Most importantly, the Channel Mixer parameters are shared across all OD pairs, implicitly fostering temporal interactions between distinct OD pairs without necessitating an increase in parameter count. This shared parameter design contributes to the model's efficiency in learning global temporal patterns, thereby enhancing its capacity to comprehend and predict variations in traffic over time.

\subsubsection{Multi-view Mixer}
The diverse relations among different OD pairs, such as the similarity in traffic variations among OD pairs connecting similar types of stations, hold the potential to enhance the accuracy of OD predictions. However, calculating the relations among $N^2$ OD pairs using attention models, such as Transformer, demands a considerable amount of time. Moreover, the direct calculation of relations among $N^2$ OD pairs may pose challenges in discerning the relations crucial for actual prediction. Therefore, efficiently computing the relations among OD pairs is a crucial and challenging problem.

To address this issue, we introduce the Multi-view Mixer module by analyzing relations among OD pairs from multiple perspectives. Recognizing that each OD pair represents the flow between two stations, which is correlated with the origin and destination, we propose the Multi-view Mixer module, consisting of two distinct parts: Origin Mixer and Des Mixer.

\textbf{Origin Mixer Module}: From the perspective of the origin within an OD pair, this module evaluates all OD pairs sharing the same origin. Concretely, we apply a dimensional transformation to the input feature $\boldsymbol{H^C}$ to obtain the transformed feature $\boldsymbol{H^{C_O}} \in \mathbb{R}^{(N \times d) \times N}$, where the first $N$ is the number of origins, the second $N$ is the number of destinations. Then we calculate the relations among OD pairs as indicated by:
\begin{equation}
\begin{aligned}
\boldsymbol{H^{O}_{i}} &= \boldsymbol{W^{O_2}}\sigma(\boldsymbol{W^{O_1}}\boldsymbol{H^{C_O}_{i}})
\end{aligned}
\end{equation}
where $i = 1, \cdots, N \times d$, $\boldsymbol{W^{O_1}} \in \mathbb{R}^{d_O \times N}$ and $\boldsymbol{W^{O_2}} \in \mathbb{R}^{N \times d_O}$ are the learnable weights of the Origin Mixer, $H_{O} \in \mathbb{R}^{N \times N \times d}$ denotes the output of the module.

\textbf{Des Mixer Module}: Analogous to the Origin Mixer, this module concentrates on relations among OD pairs from the destination perspective. To be more specific, we initially interchange the two dimensions of origin and destination, and subsequently transform the dimensions in the same way as the Origin Mixer to derive $\boldsymbol{H^{C_D}} \in \mathbb{R}^{(N \times d) \times N}$, where the first $N$ is the number of destinations and the second $N$ is the number of origins. The calculation is:
\begin{equation}
\begin{aligned}
\boldsymbol{H^{D}_{i}} &= \boldsymbol{W^{D_2}}\sigma(\boldsymbol{W^{D_1}}\boldsymbol{H^{C_D}_{i}}) 
\end{aligned}
\end{equation}
Where, $i = 1, \cdots, N \times d$, $\boldsymbol{W^{D_1}} \in \mathbb{R}^{d_D \times N}$ and $\boldsymbol{W^{D_2}} \in \mathbb{R}^{N \times d_D}$ are the learnable weights of the Des Mixer, $\boldsymbol{H^{D}} \in \mathbb{R}^{N \times N \times d}$ is the output of the Des Mixer.

Then, to obtain features after comprehensive interaction with the remaining OD pairs, a fusion approach is employed to maintain simplicity while reducing computation time:
\begin{equation}
\boldsymbol{H^F} = LayerNorm(\boldsymbol{H^O} + \boldsymbol{H^D} + \boldsymbol{H^C})
\end{equation}
where $\boldsymbol{H^F} \in \mathbb{R}^{N \times N \times d}$.

This design enables more efficient modeling of relations among complex OD pairs, thereby enhancing the model's accuracy in OD prediction. Simultaneously, by considering relations from different perspectives, the model comprehensively captures the intricate dynamics among stations.

\subsection{Bidirectional Trend Learner (BTL)}

Due to the regularity observed in people's daily routines, metro traffic exhibits temporal similarities. To effectively leverage this temporal consistency and provide a reference for the trend in OD traffic changes at the current time, we introduce a BTL. This learner discerns the temporal similarities and change trends in the flow in two directions: from the past to the present and from the present to the past, respectively.

The bidirectional trend learner comprises two symmetric components, one of which is delineated below. Initially, we aggregate two temporal features and subsequently interact these features using 1D convolution to obtain the feature $H_{f}$, representing the perceptual flow features over time:
\begin{equation}
\boldsymbol{H_{f}} = \mathrm{Conv1d}([\boldsymbol{H_{\mathrm{prev}}}; \boldsymbol{H_{\mathrm{cur}}}])
\end{equation}
$[;]$ denotes the concatenation of features, $\boldsymbol{H_{\mathrm{prev}}}, \boldsymbol{H_{\mathrm{cur}}} \in \mathbb{R}^{N \times N \times d}$ represent the features of two branches, respectively, $\boldsymbol{H_{f}} \in \mathbb{R}^{N \times N \times d}$ is the feature after the interaction.

Subsequently, the impact of the perceived change on $\boldsymbol{H_{\mathrm{prev}}}$, denoted as $\boldsymbol{H_{g}} \in \mathbb{R}^{N \times N \times d}$, is obtained through the Sigmoid activation function:
\begin{equation}
\boldsymbol{H_g} = \delta(\boldsymbol{W^g}\boldsymbol{H_f})
\end{equation}
where $\boldsymbol{W^g} \in \mathbb{R}^{d \times d}$ is the learnable weights, $\delta$ denotes the Sigmoid activation function.

Finally, we apply this effect to $\boldsymbol{H_{\mathrm{prev}}}$ to obtain $\boldsymbol{\tilde{H}_{\mathrm{prev}}} \in \mathbb{R}^{N \times N \times d}$, capturing the trend of the flow from the present to the past:
\begin{equation}
\boldsymbol{\tilde{H}_{\mathrm{prev}}} = \boldsymbol{H_{g}} \ast (\boldsymbol{W^p}\boldsymbol{H_{\mathrm{prev}}}) + \boldsymbol{H_{\mathrm{prev}}}
\end{equation}
where $\boldsymbol{W^p} \in \mathbb{R}^{d \times d}$, $\ast$ is the Hadamard product.

With the bidirectional trend learner, we derive features that perceive the traffic trend in both directions over time:
\begin{equation}
\boldsymbol{\tilde{H}_{\mathrm{prev}}}, \boldsymbol{\tilde{H}_{\mathrm{cur}}} = \mathrm{BTL}(\boldsymbol{H_{\mathrm{prev}}}, \boldsymbol{H_{\mathrm{cur}}})
\end{equation}

This design empowers the model to comprehensively grasp temporal similarities and trends in both the past-to-present and present-to-past directions, thereby enhancing the accuracy of OD flow predictions at the current time.

\subsection{Optimization}
In this subsection, we first introduce double-branch architeture of the model and then proceed to detail the model's loss function. Subsequently, we perform time complexity analysis for each component and the model.
\subsubsection{Loss Function}
To fully exploit the temporal relations within OD traffic, we establish two branches to model the past and present variations in OD traffic. Here, `prev` represents the OD traffic at the same time yesterday, and `cur` represents the OD traffic at the current time. Notably, the layer parameters of these two branches are shared. We formulated two loss functions to simultaneously optimize both branches, thereby further enriching the temporal attributes embedded in the learned features.

Ultimately, the loss function of the ODMixer is defined as:
\begin{equation}
\begin{aligned}
\mathrm{Loss} &= L_{prev} + L_{cur} \\
\red{L_{prev}} &\red{=} \red{\frac{\sum_{i=1}^{N}\sum_{j=1}^{N}|\widetilde{OD}_{t+1}^{d-1}(i, j) - OD_{t+1}^{d-1}(i, j)|}{N^{2}}} \\
\red{L_{cur}} &\red{=} \red{\frac{\sum_{i=1}^{N}\sum_{j=1}^{N}|\widetilde{OD}_{t+1}^{d}(i, j) - OD_{t+1}^{d}(i, j)|}{N^{2}}} \\
\end{aligned}
\end{equation}
where $\widetilde{OD}_{t+1}^{d-1}$, $\widetilde{OD}_{t+1}^{d}$ and $OD_{t+1}^{d-1}$, $OD_{t+1}^{d}$ represent the predicted and true values of the OD matrix at the $(t+1)^{th}$ time interval of $(d-1)^{th}$ or $d^{th}$ day, respectively.

This loss function is designed to prompt the model to learn trends in both yesterday's and today's OD traffic, facilitating a more robust capture of temporal relations. Moreover, this approach aims to enhance the accuracy of OD forecasts by better understanding and predicting the evolving patterns of OD traffic over time.

\subsubsection{Model Complexity}
The ODMixer model, comprising two fundamental components, namely ODIM and BTL, is analyzed for its computational complexities.

ODIM consists of two components: Channel Mixer and Multi-view Mixer. For Channel Mixer, the computational cost primarily arises from matrix multiplication in the linear layer, with intermediate dimensions being $2d$. This results in approximately $4N^{2}d^{2}$ operations. Since $d$ is chosen independent of $N$, the time complexity of the Channel Mixer is $O(N^2)$. 

The Multi-view Mixer module comprises two components: the Origin Mixer and the Des Mixer, with both components sharing a similar structure. Taking Origin Mixer as an example, the number of operations in the module is around $4N^{2}d^{2}$, resulting in a time complexity of $O(N^2)$. 

For the BTL module, it contains two symmetric parts. Focusing on the top half, it comprises a 1D convolution layer followed by two linear layers. The computation required for the 1D convolution is $2N^{2}d$ operations, and each linear layer requires $N^{2}d^{2}$ operations. Consequently, the time complexity of the BTL module is $O(N^{2}$).
 
Combining these analyses, the ODMixer model demonstrates a time complexity of $O(N^2)$. The training and inference time of both the baseline and ODMixer models are evaluated through experiments, as showed in Table \ref{table:hz_efficiency} and \ref{table:sh_efficiency}.

\section{Experiment}
In this section, we validate the effectiveness of our model across two extensive datasets. First, we describe the experimental configuration, including construction of the datasets, implementation details and evaluation metrics. Subsequently, we compare our proposed ODMixer with both basic and state-of-the-art models. We then conduct ablation studies to verify the influence of each component on the model's performance. Finally, we also analyse the hyper-parameter settings and the efficiency of the model.

\subsection{Experimental Configuration}
\subsubsection{Dataset Description}
We conduct experiments on two real-world datasets \cite{liu2022online}, as shown in Table \ref{table:dataset}. HZMOD records transaction data for 80 stations from January 1, 2019, to January 25, 2019. SHMOD records transaction data for 288 stations from July 1, 2016, to September 30, 2016. These transaction data are segmented into 15-minute interval, and the complete dataset is partitioned into training, validation, and test set following a specific ratio. Compared with previous subway datasets, these two datasets contain billions of transaction records from the Shanghai and Hangzhou subway systems in China, fully reflecting various real-world scenarios. The SHMOD dataset, in particular, includes 288 stations, resulting in 288*288 OD pairs, a substantial number. This scale introduces significant differences in the OD matrix, making it an excellent benchmark for assessing the model's ability to handle large and complex OD matrices.

\begin{table}[t]
\caption{Details of the datasets.}
\vspace{-5pt}
\centering
\begin{tabular}{ccc}
\toprule
Dataset & HZMOD & SHMOD \\
\midrule
Region & Hangzhou & Shanghai \\
Stations & 80 & 288 \\
OD Pairs & 80*80 & 288*288 \\
Time Interval & 15 minutes & 15 minutes \\
Training Set & 1/1/2019 - 1/18/2019 & 7/1/2016 - 8/31/2016 \\
Validation Set & 1/19/2019 - 1/20/2019 & 9/1/2016 - 9/9/2016 \\
Testing Set & 1/21/2019 - 1/25/2019 & 9/10/2016 - 9/30/2016 \\
\bottomrule
\end{tabular}
\vspace{-10pt}
\label{table:dataset}
\end{table}

\subsubsection{Implementation Details}
We implement our model in PyTorch and conduct both training and testing using 8 RTX 2080Ti GPUs. The initial learning rate for the model is set at 0.001. To optimize the model loss, we employ the Adam optimizer. The batch size is set to 32 for both datasets. The input sequence T is set to 4. The input data and the ground-truth of output are normalized with Z-score normalization before fed into the model. The feature dimension $d$ is 16 and ODMixer layer $L$ is 5. And we set $d_C = d_O = d_D = 2 \times d$.

\subsubsection{Quantitative Evaluation Metrics}
We have selected three widely employed performance metrics within the domain of traffic flow prediction for evaluating the model's effectiveness. These metrics include Mean Absolute Error (MAE), Root Mean Squared Error (RMSE), and Weighted Mean Absolute Percentage Error (wMAPE). MAE is simple and intuitive, insensitive to outliers, suitable for scenarios that require robustness. RMSE penalizes large errors, provides information about the distribution of prediction errors, and is suitable for scenarios that need to emphasize the severity of errors. wMAPE expresses errors in percentage form, suitable for comparison of datasets of different sizes, and can be weighted to reflect the importance of data points.
\begin{equation}
\begin{aligned}
\mathrm{MAE} &= \frac{1}{N^{2}}\sum_{i=1}^{N}\sum_{j=1}^{N}|\widetilde{OD}(i, j) - OD(i, j)| \\
\mathrm{RMSE} &= \sqrt{\frac{1}{N^{2}}\sum_{i=1}^{N}\sum_{j=1}^{N}(\widetilde{OD}(i, j) - OD(i, j))^{2}} \\
\mathrm{wMAPE} &= \frac{\sum_{i=1}^{N}\sum_{j=1}^{N}|\widetilde{OD}(i, j) - OD(i, j)|}{\sum_{i=1}^{N}\sum_{j=1}^{N}OD(i, j)}
\end{aligned}
\end{equation}
\subsection{Comparison with State-of-the-Art Methods}
We conduct a comparative analysis between our proposed model and a selection of classical as well as contemporary models.
STCNN \cite{zhang2023deep} and ST-VGCN\cite{yang2022spatiotemporal} that consider OD pairs aren't included because they have no publicly accessible codes and have not been evaluated on identical datasets.

\begin{itemize}
\item \textbf{Historical Average (HA):} Future projections were derived by directly utilizing historical flow averages for each metro station.
\item \textbf{LSTM\cite{memory2010long}:} This model utilizes a Seq2Seq approach, employing two LSTM layers to predict future metro OD.
\item \textbf{GRU\cite{cho2014learning}:} The architecture of this model closely resembles the previous one, with a key distinction being the use of GRU layers instead of LSTM layers.
\item \textbf{Graph WaveNet\cite{wu2019graph}:} The model acquires adaptive dependency matrices within the data by embedding nodes, and it captures long-range temporal dependencies through the use of stacked dilated 1D convolutions.
\item \textbf{DCRNN\cite{li2018diffusion}:} DCRNN captures spatial dependencies through the utilization of bi-directional random walks on the graph. Additionally, it employs encoder-decoder structures with scheduled sampling to model temporal information.
\item \textbf{STG2Seq\cite{bai2019stg2seq}:} STG2Seq incorporates a hierarchical graph convolution structure to effectively capture both spatial and temporal correlations.
\item \textbf{PVCGN\cite{liu2020physical}:} PVCGN comprehensively learns complex spatio-temporal relations between metro stations by constructing physical, similarity, and correlation graphs.
\item \textbf{DGSL\cite{shang2021discrete}:} DGSL optimizes the average performance of the graph distribution by leveraging probabilistic graph models when the graph structure is unknown, thus facilitating the learning of the graph structure.
\item \textbf{Informer\cite{zhou2021informer}:} Informer is a transformer-based model specifically designed for the task of long sequence time-series forecasting. It is notable for its efficient ProbSparse self-attention mechanism, self-attention distillation, and generative-style decoder. For our metro OD prediction, we built our model using the official code of Informer as a foundation, and we customized it to suit our specific requirements.
\item \textbf{HIAM\cite{liu2022online}:} HIAM integrates diverse information to jointly learn the evolutionary patterns of OD and OD, featuring two branch for OD and DO estimation separately and a DIT for enhanced OD-DO correlation modeling.
\red{\item \textbf{HSTN\cite{chen2022origin}:} HSTN designs a HSM to capture three types of spatial relations and a HTM to quantify the influence of the input sequence on the target result.}
\red{\item \textbf{MVPF\cite{zheng2022metro}:} MVPF considers multiple views of real-time traffic information including individual station and cross-station flow information.}
\red{\item \textbf{C-AHGCSP\cite{ye2024heterogeneous}:} C-AHGCSP completes unfinished trips based on real-time mobility evolution and inter-station time costs to enhance dynamic adaptability and considers the destination distributions of the passengers departing from a station are correlated with other stations sharing similar attributes.}
\end{itemize}

\begin{table}[!t]\small
\caption{Performance of OD prediction on HZMOD dataset.}
\vspace{-5pt}
\centering
\begin{tabular}{cccc}
\toprule
Models & MAE ($\downarrow$) & RMSE ($\downarrow$) & wMAPE ($\downarrow$)\\
\midrule
HA & 1.355 & 2.917 & 48.354\% \\
LSTM & 1.387 & 3.458 & 49.500\% \\
GRU & 1.427 & 3.593 & 50.906\% \\
Graph WaveNet & 1.717 & 4.431 & 62.489\% \\
\red{HSTN} & \red{1.692} & \red{5.558} & \red{60.392\%} \\
\red{MVPF} & \red{1.805} & \red{4.749} & \red{64.414\%} \\
DCRNN & 1.269 & 2.944 & 46.203\% \\
STG2Seq & 1.578 & 4.355 & 56.302\% \\
DGSL & 1.244 & 2.906 & 45.269\% \\
Informer & 1.272 & 2.756 & 45.393\% \\
PVCGN & 1.241 & 2.697 & 44.290\% \\
\red{C-AHGCSP} & \red{1.239} & \red{2.694} & \red{44.200\%} \\
HIAM & \underline{1.196} & \underline{2.581} & \underline{42.690\%} \\
\textbf{ODMixer (Ours)} & \textbf{1.131} & \textbf{2.367} & \textbf{40.358\%} \\
\bottomrule
\end{tabular}
\vspace{-10pt}
\label{table:hz_performance}
\end{table}

\begin{table}[!t]\small
\caption{Performance of OD prediction on SHMOD dataset.}
\vspace{-5pt}
\centering
\begin{tabular}{cccc}
\toprule
Models & MAE ($\downarrow$) & RMSE ($\downarrow$) & wMAPE ($\downarrow$) \\
\midrule
HA & 0.515 & 1.429 & 70.388\% \\
LSTM & 0.507 & 1.789 & 69.261\% \\
GRU & 0.526 & 1.920 & 71.887\% \\
Graph WaveNet & 0.560 & 2.049 & 76.851\% \\
\red{HSTN} & \red{0.560} & \red{2.530} & \red{76.520\%} \\
\red{MVPF} & \red{0.587} & \red{2.271} & \red{80.185\%} \\
DCRNN & 0.442 & 1.392 & 61.357\% \\
STG2Seq & 0.651 & 2.703 & 88.887\% \\
DGSL & \underline{0.431} & 1.238 & \underline{59.820\%} \\
Informer & 0.482 & \underline{1.118} & 65.883\% \\
PVCGN & 0.441 & 1.229 & 60.184\% \\
\red{C-AHGCSP} & \red{0.564} & \red{2.096} & \red{77.039\%} \\
HIAM & 0.441 & 1.226 & 60.268\% \\
\textbf{ODMixer (Ours)}  & \textbf{0.408} & \textbf{1.096} & \textbf{55.691\%} \\
\bottomrule
\end{tabular}
\vspace{-10pt}
\label{table:sh_performance}
\end{table} 

\begin{figure}[!t]
    \centering
    \subfloat[Impact of Number of Layer]{
        \includegraphics[width=85mm]{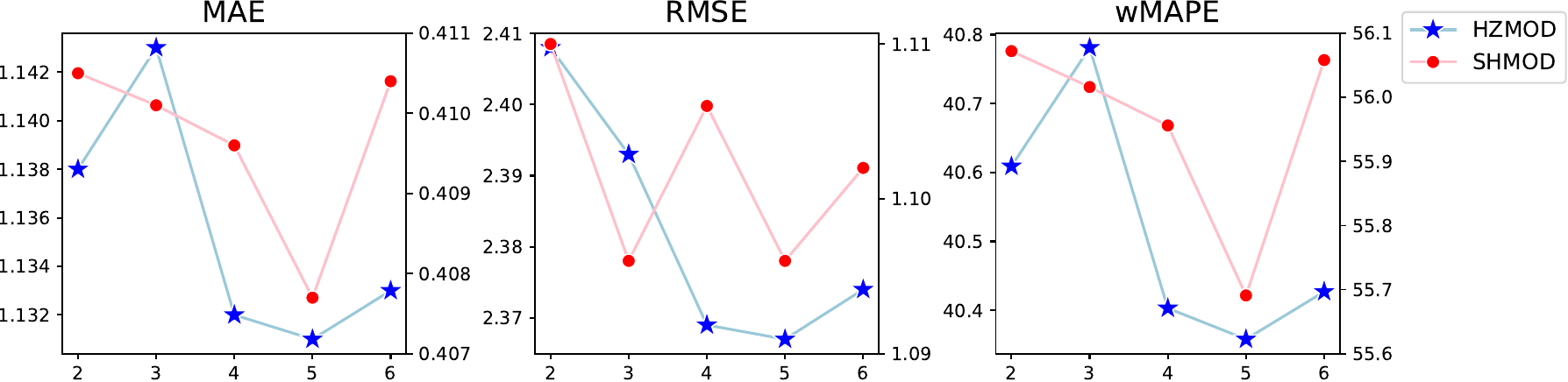}
    }
    \\
    \subfloat[Impact of Number of Dimension]{
        \includegraphics[width=85mm]{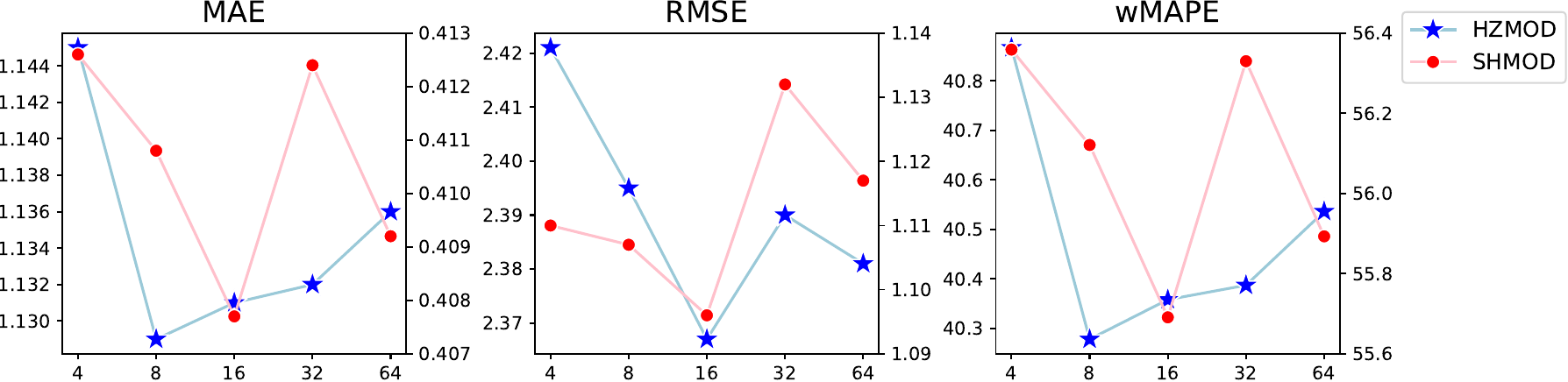}
    }
    \vspace{-5pt}
    \caption{Impact of Hyper-parameters. The two figure illustrate the impact of different parameters on prediction results across two datasets. Figure (a) displays the effect of varying the number of layers on performance, while Figure (b) shows the influence of different feature dimensions on performance.}
    \vspace{-10pt}
\label{fig:hyper}
\end{figure}

We have conducted experiments on two datasets, and the experimental results for all methods are presented in Table \ref{table:hz_performance} and \ref{table:sh_performance}. \red{We use the same data inputs as specified in the original papers when evaluating all baselines, provided the relevant data is available. For example, HIAM utilizes additional information, such as the Destination-Origin (DO) matrix, which we incorporate in our experiments.} Firstly, the simplest method, HA, can partially reflect the flow range of the training set by directly averaging it. However, lacking generalization ability, it fails to capture temporal variations in traffic. Consequently, its performance tends to be inferior to that of ODMixer, with the performance gap widening as the dataset size increases. \red{After analysis, it is evident that the performance of LSTM, GRU, Graph WaveNet, STG2Seq, HSTM and MVPF models is inferior to that of HA, potentially due to their oversimplified design that neglects the spatial-temporal dependencies among metro stations. C-AHGCSP addresses the issue of incomplete OD matrices by introducing a dedicated module for completion. However, its approach involves sequentially filling in each time interval step by step. And it primarily utilizes the flow information from the most recent moment and uses simple module to learn dependencies. Therefore it performs well in smaller-scale metro networks, it faces limitations when dealing with larger-scale metro networks} The incorporation of spatial-temporal relations in models like DCRNN and Informer has resulted in better performance. DGSL outperforms other baselines on the SHMOD dataset because it learns the graph structure rather than using the provided structure, which allows for a more effective capture of relations on complex dataset. PVCGN, which considers three types of graphs and comprehensively accounts for the inter-station relations, exhibits good performance across both datasets. HIAM, which integrates OD and DO information and considers the incompleteness of OD matrices, achieves commendable results but performs less effectively than DGSL on the SHMOD dataset, possibly because the large scale of SHMOD requires a more fine-grained approach than can be provided by neighborhood graph aggregation alone to capture the complex interactions among numerous stations.

\begin{table}[!t]\small
\caption{Performance of different variants on HZMOD dataset.}
\vspace{-5pt}
\centering
\begin{tabular}{cccc}
\toprule
Variants & MAE ($\downarrow$) & RMSE ($\downarrow$) & wMAPE ($\downarrow$)\\
\midrule
ODMixer-w/o OMP & 1.159 & 2.487 & 41.366\% \\
ODMixer-w/o CM & 1.138 & 2.390 & 40.614\% \\
ODMixer-w/o OM & 1.149 & 2.409 & 41.015\% \\
ODMixer-w/o DM & 1.154 & 2.428 & 41.177\% \\
ODMixer-w/o MM & 1.196 & 2.534 & 42.675\% \\
ODMixer-w/o BTL & 1.132 & 2.375 & 40.397\% \\
ODMixer-w/o PB & 1.137 & 2.388 & 40.581\% \\
\textbf{ODMixer (Ours)}  & \textbf{1.131} & \textbf{2.367} & \textbf{40.358\%} \\
\bottomrule
\end{tabular}
\vspace{-15pt}
\label{table:hz_ablation}
\end{table}

In contrast, ODMixer models the OD matrix with fine granularity from the perspective of OD pairs. Subsequently, it leverages spatial-temporal relations among OD pairs through the ODIM module and integrates long-term temporal dynamics with the BTL module, thereby achieving superior performance.

\begin{table}[!t]\small
\caption{Performance of different variants on SHMOD Dataset.}
\vspace{-5pt}
\centering
\begin{tabular}{cccc}
\toprule
Variants & MAE ($\downarrow$) & RMSE ($\downarrow$) & wMAPE ($\downarrow$)\\
\midrule
ODMixer-w/o OMP & 0.418 & 1.126 & 57.064\% \\
ODMixer-w/o CM & 0.410 & 1.107 & 56.039\% \\
ODMixer-w/o OM & 0.414 & 1.133 & 56.602\% \\
ODMixer-w/o DM & 0.416 & 1.126 & 56.862\% \\
ODMixer-w/o MM & 0.420 & 1.152 & 57.319\% \\
ODMixer-w/o BTL & 0.410 & \textbf{1.090} & 55.979\% \\
ODMixer-w/o PB & 0.412 & 1.100 & 56.257\% \\
\textbf{ODMixer (Ours)}  & \textbf{0.408} & 1.096 & \textbf{55.691\%} \\
\bottomrule
\end{tabular}
\vspace{-10pt}
\label{table:sh_ablation}
\end{table}

\subsection{Ablation Study}



In this subsection, we aim to validate the effectiveness of essential modules within ODMixer. By systematically removing each module, we derive variants of ODMixer. The subsequent comparison of performance differences among these variants and the original ODMixer allows us to understand the specific contributions of each module to the overall model. The following variants will be evaluated on two datasets:
\begin{itemize}
    \item \textbf{ODMixer-w/o OMP:} Remove OD Matrix Processing module, eliminating the impact of unfinished orders.
    \item \textbf{ODMixer-w/o CM:} Remove Channel Mixer, eliminating short-term temporal relations modeling.
    \item \textbf{ODMixer-w/o OM:} Remove Origin Mixer, eliminating the Origin perspective in OD pair relations modeling.
    \item \textbf{ODMixer-w/o DM:} Remove Des Mixer, eliminating the Destination perspective in OD pair relations modeling.
    \item \textbf{ODMixer-w/o MM:} Remove Multi-view Mixer, eliminating relations modeling among OD pairs.
    \item \textbf{ODMixer-w/o BTL:} Remove BTL, eliminating the perception of long-term traffic changes.
    \item \textbf{ODMixer-w/o PB:} Remove Prev Branch and BTL, eliminating the usage of historical information.
\end{itemize}

The results are presented in the Table \ref{table:hz_ablation} and \ref{table:sh_ablation}. ODMixer-w/o OMP removes the OD Matrix Processing module, the Cur Branch receives only the incomplete OD matrix as input, without any completion processing. This incompleteness results in the model lacking critical information needed to capture the traffic characteristics at the current time, leading to an inaccurate reflection of the actual traffic situation. Consequently, the model struggles to make effective predictions, and its performance drops significantly. ODMixer-w/o CM exhibits inferior performance compared to ODMixer, signifying the effectiveness of Channel Mixer in modeling short-term temporal relations of OD pairs. ODMixer-w/o MM performs less effectively than both ODMixer-w/o OM and ODMixer-w/o DM, highlighting the necessity of both Origin and Destination perspectives. The results indicate that these two perspectives complement each other in considering the relations among OD pairs. \red{The Multi-view Mixer's effectiveness can be attributed to the fact that each OD pair is associated with both an origin and a destination, and the attributes of these locations influence the passenger flow dynamics. For example: If the origin is a residential area and the destination is a business district, there is typically a morning peak in passenger flow, with lower flows during other times. These patterns validate the effectiveness and rationale of modeling spatial relations from both dimensions.}  ODMixer-w/o BTL and ODMixer-w/o PB show some performance losses, suggesting that accounting for the temporal similarity of metro flows can enhance the model's performance. Due to the difficulty of accurately modeling long-term dependencies for large-scale metro data, ODMixer performs slightly worse than ODMixer-w/o BTL in RMSE metrics. These experimental findings affirm that all modules contribute to the ODMixer.

\subsection{Hyper-parameters Analysis}
We conduct experiments to optimize the hyperparameters of ODMixer, with the results depicted in Figure \ref{fig:hyper}. The hyperparameters for ODMixer include the number of layer $L$ and the number of channel $d$. Notably, when $L$ is small, the model exhibits weak feature extraction capabilities, resulting in comparatively inferior performance. Conversely, a large $L$ introduces fluctuations in performance. After comprehensive experimentation, we establish the optimal setting as $L=5$. Regarding the dimension of the hidden layer, experimental findings indicate that $d=16$ is a better choice for achieving desirable model performance.

\begin{table}[!t]\small\setlength{\tabcolsep}{0.8mm}
\caption{Efficiency study of ODMixer on HZMOD dataset.}
\vspace{-5pt}
\centering
\begin{threeparttable}
\begin{tabular}{ccccc}
\toprule
Models & MAE & Parameters (M) & Train (s) & Infer (s) \\
\midrule
LSTM & 1.387 & 1.77& 2.14 & 0.30 \\
GRU & 1.427 & 1.33 & 2.18 & 0.31 \\
Graph WaveNet & 1.717 &1.21 & 1.82 & 0.16 \\
\red{HSTN} & \red{1.692} & \red{2.18} & \red{6.51} & \red{1.86} \\
\red{MVPF} & \red{1.805} & \red{2.51} & \red{3.55} & \red{1.66} \\
DCRNN & 1.269 & 6.54 & 13.40 & 1.93 \\
STG2Seq & 1.578 & 1.06 & 2.17 & 0.23 \\
DGSL & 1.244 & 8.55 & 12.44 & 1.78 \\
Informer & 1.272 & 88.69 & 2.21 & 2.76 \\
PVCGN$^{*}$ & 1.241 & 55.84 & 94.11 & 10.43 \\
\red{C-AHGCSP} & \red{1.239} & \red{2.17} & \red{3.98} & \red{1.72} \\
HIAM & 1.196 & 13.89 & 13.21 & 1.25 \\
\textbf{ODMixer (Ours)} & \textbf{1.131} & 2.12 & 5.61 & 1.18 \\
\bottomrule
\end{tabular}
\begin{tablenotes}
\footnotesize
\item[$*$] The Batch Size in PVCGN is 16.
\end{tablenotes}
\end{threeparttable}
\vspace{-10pt}
\label{table:hz_efficiency}
\end{table}

\subsection{Efficiency Analysis}

\begin{table}[!t]\small\setlength{\tabcolsep}{0.8mm}
\caption{Efficiency study of ODMixer on SHMOD dataset.}
\vspace{-5pt}
\centering
\begin{tabular}{ccccc}
\toprule
Models & MAE & Parameters (M) & Train (s) & Infer (s) \\
\midrule
LSTM & 0.507 & 2.24 & 55.44 & 9.91 \\
GRU & 0.526 & 1.70 & 52.19 & 10.48 \\
Graph WaveNet & 0.560 & 1.44 & 43.74 & 3.88 \\
\red{HSTN} & \red{0.560} & \red{6.98} & \red{80.81} & \red{10.93} \\
\red{MVPF} & \red{0.587} & \red{3.81} & \red{40.22} & \red{8.67} \\
DCRNN & 0.442 & 28.17 & 165.19 & 27.55 \\
STG2Seq & 0.651 & 4.07 & 27.66 & 4.48 \\
DGSL & 0.431 & 35.33 & 240.37 & 39.76 \\
Informer & 0.482 & 637.43 & - & - \\
PVCGN & 0.441 & 178.15 & 640.56 & 90.72 \\
\red{C-AHGCSP} & \red{0.564} & \red{2.31} & \red{61.36} & \red{14.01} \\
HIAM & 0.441 & 28.01 & 286.68 & 36.19 \\
\textbf{ODMixer (Ours)}  & \textbf{0.408} & 26.75 & 247.91 & 26.86 \\
\bottomrule
\end{tabular}
\label{table:sh_efficiency}
\end{table}

In this subsection, we quantify the number of parameters for the various models. To ensure a fair comparison, all models are evaluated on a single NVIDIA GeForce RTX 2080 Ti GPU with a consistent batch size. The batch size is 8 for SHMOD and 32 for HZMOD, respectively.

Table \ref{table:hz_efficiency} and \ref{table:sh_efficiency} present a comparative analysis of all methods. \red{The LSTM, GRU, Graph WaveNet, STG2Seq, HSTN and MVPF models exhibit poorer performance than HA, despite their shorter training and inference times. Because C-AHGCSP completes the OD matrix step by step introduces additional computational overhead, it needs longer inference times.} Despite Informer's short training and inference times on the HZMOD dataset, it has a substantial number of parameters. Regarding the SHMOD dataset, Informer cannot be trained , even when the batch size is 1. Furthermore, the PVCGN model is unable to be trained on a single GPU with a batch size of 32 for HZMOD, necessitating a reduction to 16 for testing. Despite its good performance, the PVCGN model requires significantly longer training and inference times. In contrast, the HIAM model, which serves as the best-performance baseline, has a higher number of parameters, as well as longer training and inference times compared to ODMixer. Our proposed ODMixer not only achieves superior performance but also maintains reasonable training and inference times.

\subsection{Validity Verification}
Since our method is based on the perspective of OD pairs, while most existing methods are station-based, we aim to verify whether our model structure can effectively address the issues in existing methods from the same perspective. To this end, we replaced the ODIM module in ODMixer with a standard Attention module, naming the modified model ODTrans. We then conducted experiments, and the results are presented in the following Table \ref{table:Validity_Verification}.

\begin{table}[!t]\setlength{\tabcolsep}{0.3mm}
\caption{\red{Validity Verification on HZMOD Dataset.}}
\vspace{-5pt}
\centering
\begin{threeparttable}
\setlength{\tabcolsep}{0.7mm}\begin{tabular}{ccccccccc}
\toprule
Models & wMAPE ($\downarrow$) & Train (s) & Test (s) & Layer & Dim & BatchSize & GPUs \\
\midrule
$\text{ODTrans}^{1}$ & 47.422\% & 19.02 & 3.92 & 2 & 16 & 8 & 8 \\
ODMixer & \textbf{40.609\%} &  \red{2.85} & \red{0.88} & 2 & 16 & 32 & 1 \\
$\Delta$ & 14,37\% & \red{$\times 6.67$} & \red{$\times 4.45$} & - & - & - & - \\
\hline
$\text{ODTrans}^{1}$ & 47.007\% & 27.80 &  4.70 & 3 & 16 & 8 & 8 \\
ODMixer & \textbf{40.781\%} &  \red{3.75} & \red{0.98} & 3 & 16 & 32 & 1 \\
$\Delta$ & 13.24\% & \red{$\times 7.41$}  & \red{$\times 4.80$} & - & - & - & - \\
\bottomrule
\end{tabular}
\begin{tablenotes}
\footnotesize
\item[1] For the Transformer, we set the batch size to 8, the number of layers to 2 or 3, and use 8 GPUs to ensure proper execution.
\end{tablenotes}
\end{threeparttable}
\vspace{-10pt}
\label{table:Validity_Verification}
\end{table}

\begin{figure}[!t]
    \centering
        \subfloat[Case for SHMOD Dataset]{
        \includegraphics[width=40mm]{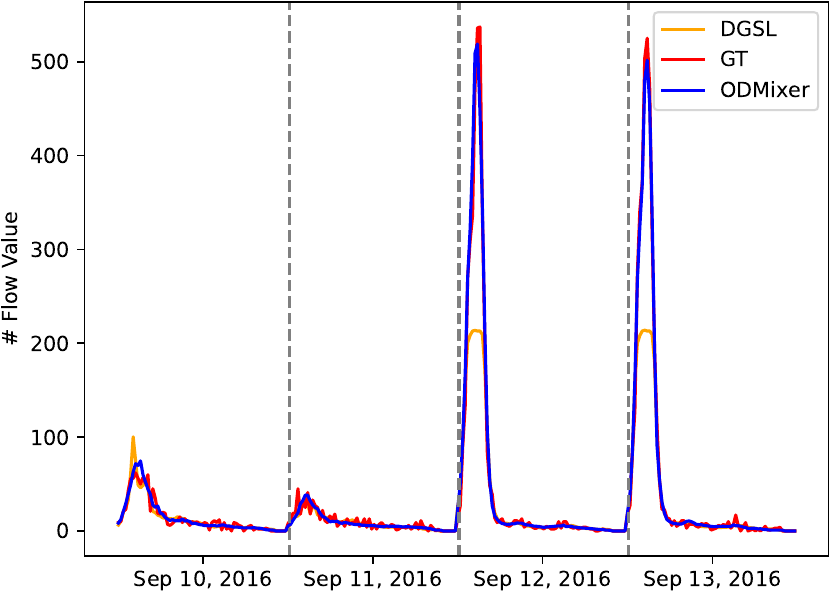}
    }
    \subfloat[Case for HZMOD Dataset]{
        \includegraphics[width=40mm]{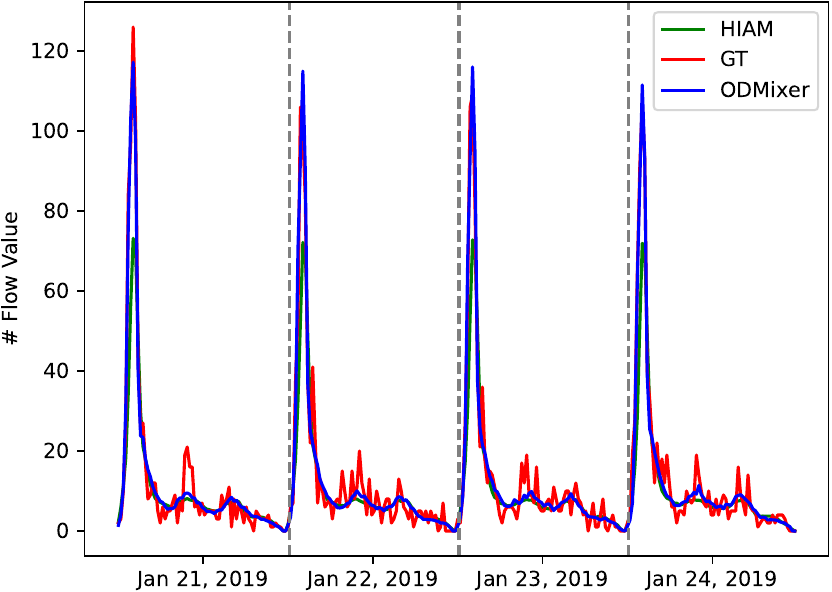}

    }
    \caption{Case for SHMOD and HZMOD Dataset. (a) shows the prediction results of the DGSL and ODMixer models from September 10 to 13, with the 12th being a Monday. (b) shows the prediction results of the HIAM and ODMixer models from January 21 to 24, with the 21th being a Monday.  It can be observed that when traffic flow changes rapidly, the performance of HIAM and DGSL is significantly worse than that of ODMixer.}
    \vspace{-10pt}
\label{fig:case}
\end{figure}

As described in the introduction, the number of input tokens is $N^2$ from the perspective of OD pairs, resulting in space and time complexities of $N^4$ for the attention mechanism. Consequently, ODTrans cannot run with the same number of GPUs and batch size as ODMixer when using the same model layers and dimensions. For a fair performance comparison, we reduced the number of model layers and batch size and employed multiple GPUs for ODTrans. We also used the same number of layers for training ODMixer to ensure fairness.

The experiments show that the performance of ODMixer is improved by 14.37\% and 13.24\% at layers 2 and 3 compared to ODTrans, respectively. Additionally, by comparing training and testing times, we found that ODMixer's training time and testing time at layer 2 were accelerated by \red{6.67} times and \red{4.45} times, respectively. At layer 3, training and testing times were accelerated by \red{7.41} times and \red{4.80} times, respectively. These results demonstrate that the ODMixer module, designed specifically for the metro OD prediction task, effectively captures the spatiotemporal relations in traffic. This significantly enhances the model's operational efficiency, reduces the required space, and greatly improves its deployability.

\subsection{Case Study}

In this subsection, we use two practical examples to demonstrate why our ODMixer achieves better results. 

As shown in Fig. \ref{fig:case} (a), the OD traffic on Saturday and Sunday (Sep 10-11) is minimal and stable, allowing both DGSL and ODMixer to fit the real traffic well. However, during the peak periods of Monday and Tuesday (Sep 12-13), the traffic between OD pairs changes rapidly. Although DGSL senses the growth and attenuation of traffic, its estimated peak value remains small. Conversely, ODMixer quickly detects these rapid changes and accurately estimates the peak value of the OD pair's traffic. This is mainly due to ODMixer's input modeling based on the OD pair perspective. The Channel Mixer captures the short-term temporal relations, the Multi-view Mixer learns the spatial relations between OD pairs, and BTL senses long-term traffic similarity. These designs enable ODMixer to handle rapid changes and complex relations in traffic effectively, significantly improving the model's prediction accuracy and adaptability. A similar situation is observed with HIAM and our ODMixer in Fig. \ref{fig:case} (b).

\red{\subsection{Scalability}}

\begin{figure}[!t]
    \centering
    \subfloat[Time curves of different L]{
        \includegraphics[width=40mm]{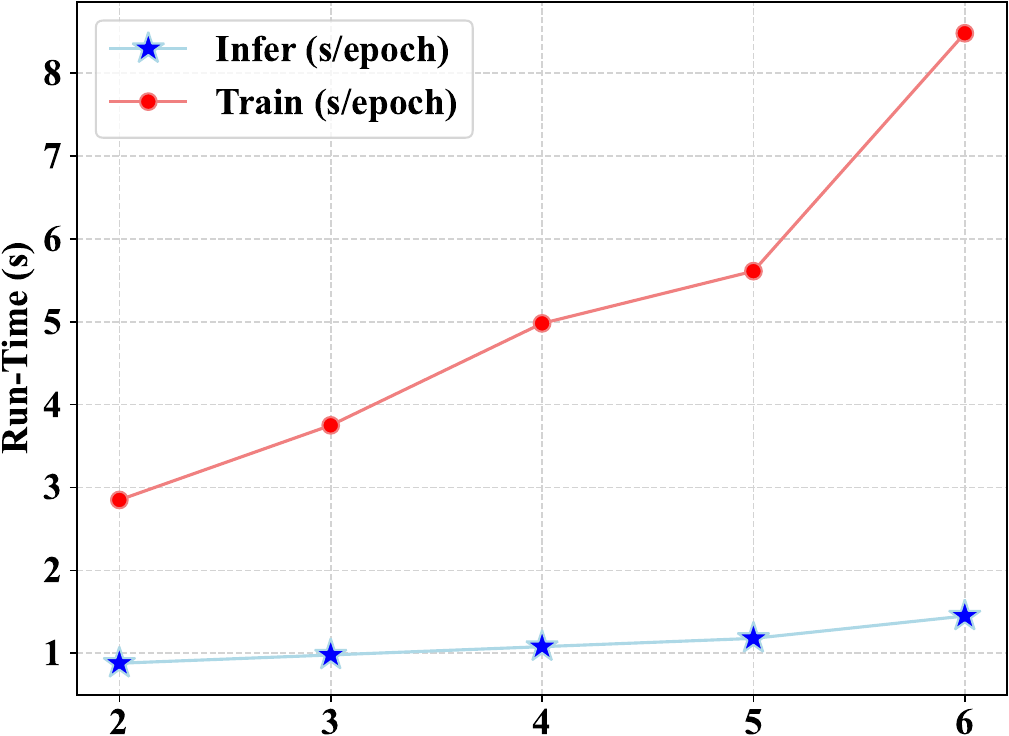}
    }
    \subfloat[Time curves of different d]{
        \includegraphics[width=40mm]{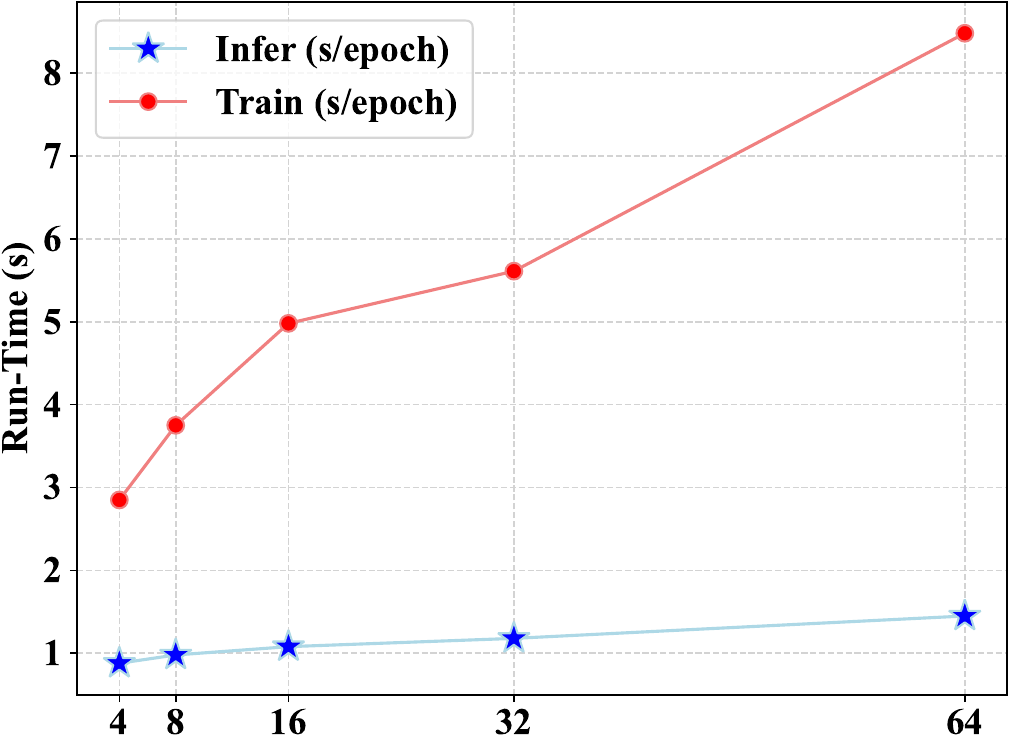}
    }
    \caption{Comparasions between running time and various L, d.}
    \vspace{-5pt}
\label{fig:scalability}
\end{figure}

\red{
Scalability tests are performed to assess the impact of various hyperparameters on model efficiency, with a primary focus on the number of layers $L$ and the hidden layer dimension $d$. First, we evaluate the effect of the number of layers. As shown in Fig. \ref{fig:scalability}(a), both training and inference times increase as the number of layers grows, although the increase in inference time is relatively smaller. Considering the trade-off between performance (Fig. \ref{fig:hyper}(a)) and computational cost, we select 5 layers to strike a balance between effectiveness and efficiency.
}

\red{
Next, we examine the effect of the hidden layer dimension, as depicted in Fig. \ref{fig:scalability}(b). The results indicate that training time varies with the hidden layer dimension, while inference time remains relatively stable. Based on overall performance metrics (Fig. \ref{fig:hyper}(b)), we choose a hidden layer dimension of 16. Overall, the model's inference time exhibits minimal variation, and training time remains within acceptable limits. These findings confirm that ODMixer can be efficiently trained on larger datasets and deployed in real-time applications. Thus, ODMixer demonstrates strong stability and adaptability under different hyperparameter settings, making it highly suitable for real-time scenarios.
}

\red{\subsection{Robustness}}

\begin{table}[!t]\setlength{\tabcolsep}{0.8mm}
\caption{Prediction error of Models on original data(1st row), Gaussian-noise polluted data(2nd row) and the relative increment ratio of the error(3rd row).}
\vspace{-8pt}
\centering
\begin{threeparttable}
\begin{tabular}{c|ccc|ccc}
\toprule
\multirow{2}*{Model} & \multicolumn{3}{c|}{HZMOD} & \multicolumn{3}{c}{SHMOD} \\
\cline{2-7}
~ & MAE & RMSE & wMAPE & MAE & RMSE & wMAPE \\
\midrule
HIAM & 1.196 & 2.581 & 42.690\% & 0.441 & 1.226 & 60.268\% \\
+ $\mathcal{N}(0, 1)$ & 1.214 & 2.638 & 43.337\% & 0.454 & 1.279 & 62.049\% \\
+$\Delta$errors & 1.51\% & 2.21\% & 1.52\% & 2.95\% & 4.32\% & 2.96\% \\
\hline
ODMixer  & 1.131 & 2.367 & 40.358\% & 0.408 & 1.096 & 55.691\% \\
+ $\mathcal{N}(0, 1)$ & 1.136 & 2.380 & 40.538\% & 0.413 & 1.122 & 56.384\% \\
+$\Delta$errors & 0.44\% & 0.55\% & 0.45\% & 1.23\% & 2.37\% & 1.24\% \\
\bottomrule
\end{tabular}
\vspace{-10pt}
\end{threeparttable}
\label{table:robustness_noise}
\end{table}

\red{
In real-world metro systems, hardware or software failures such as sensor malfunctions or network issues may result in missing order data, leading to discrepancies between the obtained traffic data and the actual data. To assess the robustness of the model under such conditions, we evaluate its performance under noisy and missing data scenarios. For the noisy data case, we add Gaussian noises into the raw data of test dataset and test the model on the noisy test dataset using the trained model\cite{jiang2023enhancing}. The results in Table \ref{table:robustness_noise} show the increasing errors of ODMixer are much less than SOTA for the noise data, verifying the robustness of ODMixer.
}

\begin{figure}[!t]
    \centering
    \subfloat[HZMOD]{
        \includegraphics[width=40mm]{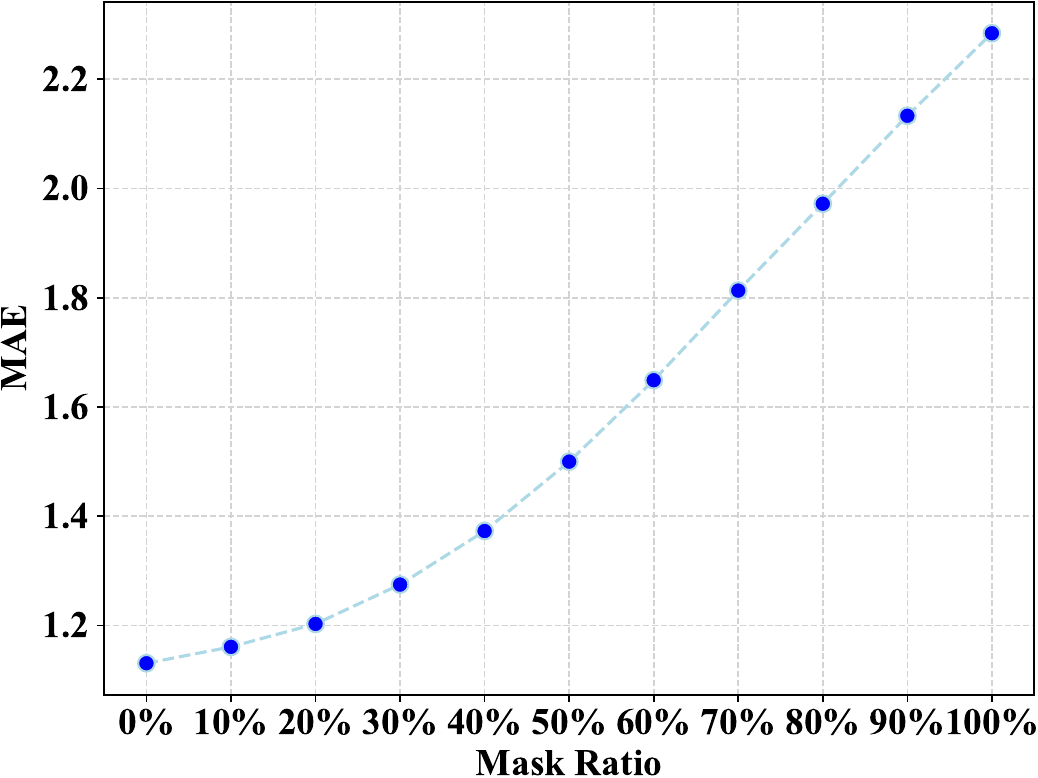}
    }
    \subfloat[SHMOD]{
        \includegraphics[width=40mm]{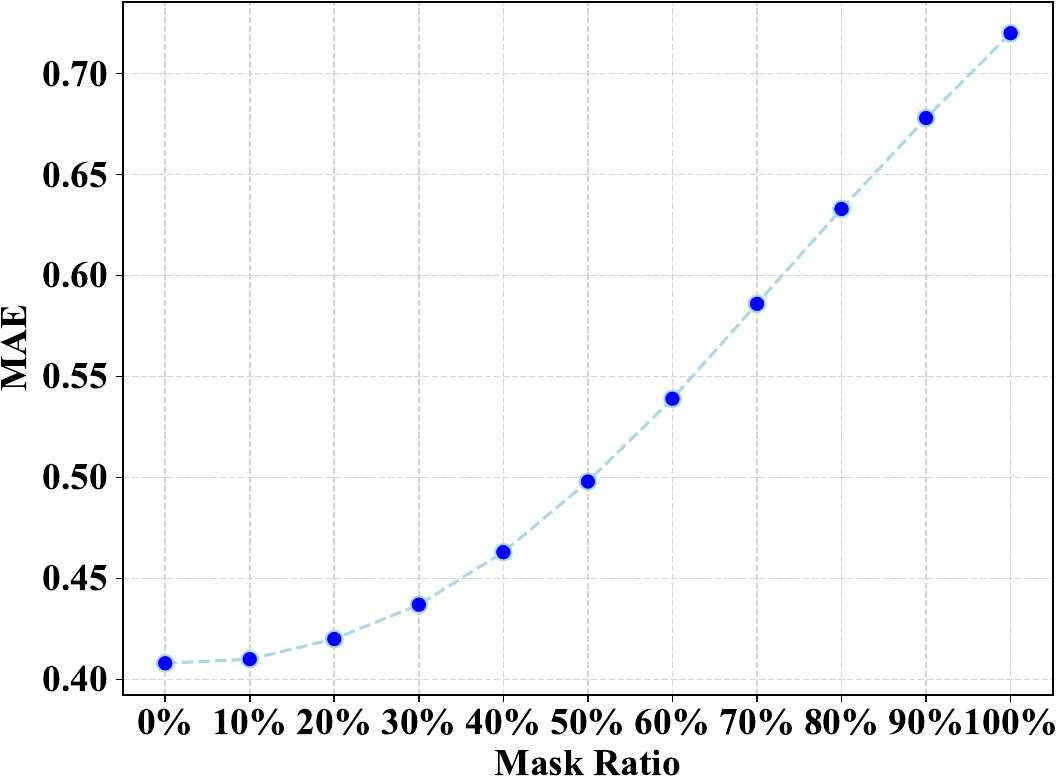}
    }
    \caption{MAE curve for different missing ratio on HZMOD and SHMOD.}
    \vspace{-15pt}
\label{fig:missing_data}
\end{figure}

\red{
For the missing data scenario, we perform spatiotemporal masking on the test dataset, randomly masking a portion of the data to simulate potential missing data situations. The results in Fig.\ref{fig:missing_data} show the prediction errors is gradually increasing when mask ratio is increasing. When the mask ratio is relatively low (not greater than $30\%$), ODMixer still maintains good performance compared to the SOTA models, demonstrating that ODMixer has some resilience to missing data. Through experiments in both noisy and missing data scenarios, we further demonstrate that ODMixer exhibits robust performance, effectively handling common issues of noise and data missing in real-world applications. This ensures the model's stability and reliablity in real-world metro environments. Traffic prediction robustness focuses on enhancing a model's resilience to noise and handling missing data, which often referred to as the imputation task. Several studies have explored model robustness \cite{jiang2023enhancing, chen2024laplacian}. Future work could integrate noise resilience and imputation modules to preprocess noisy or incomplete data, improving prediction quality.
}

\red{\subsection{Cross-City Generalizability}}

\red{
We conduct a comprehensive experiment using two real-world datasets with distinct characteristics. The results demonstrate that our model can effectively capture the complex spatiotemporal relations inherent in different datasets and is well-suited for deployment in metro systems with varying characteristics. To further validate the generalization capability of our model, we design the cross-city experiment, where the model is trained on the metro network of one city and fine-tuned and tested on a different city's network with distinct characteristics. However, due to differences in the number of stations across cities, the number of OD pairs also varies.}

\red{
Therefore, to ensure the model operates properly, we map the number of stations at the input stage accordingly. We adopted two efficient mapping methods, OAda and TAda, to address differences in the number of stations between cities. These methods use Adapters to map the number of stations from the source city to the target city ($N_{source\_city} \rightarrow N_{target\_city}$), where $N_{source\_city}$ and $ N_{target\_city}$ represent the numbers of metro stations in the source and target cities, respectively (e.g., HZMOD has 80 stations, while SHMOD has 288). Each Adapter is implemented as a two-layer MLP.
In the OAda method, the origin and destination share a single Adapter, whereas in the TAda method, the origin and destination each use separate Adapters. We test the performance of two fine-tuning approaches: full model fine-tuning and Adapter-layer-only fine-tuning. The experimental results, shown in the Table \ref{table:cross_city}, indicate that full model fine-tuning outperforms Adapter-layer-only fine-tuning and achieves better results than the current SOTA methods. Interestingly, Adapter-layer-only fine-tuning performs comparably to SOTA and even slightly surpasses it under the TAda configuration.
Comparing the two mapping methods, TAda generally outperforms OAda, especially when only the Adapter layer is fine-tuned. These results demonstrate that ODMixer has strong cross-city generalization and adaptability, effectively transferring knowledge between metro networks with different attributes.
}


\red{
Recent research has introduced methods specifically designed to enhance the generalization of traffic prediction models \cite{zhang2024personalized, mo2022cross}. These approaches aim to improve a model's ability to transfer knowledge across datasets with varying attributes. The primary objective is to train models on large, data-rich datasets and fine-tune them on smaller, domain-specific datasets, addressing challenges related to data scarcity. In future work, ODMixer could be extended to develop a more generalized metro OD prediction model.
}

\begin{table}[!t]\setlength{\tabcolsep}{0.8mm}
\caption{Cross-City performance of different methods. FF: Full Fine-tuning. OFA: Only Fine-tuning Adapter. OAda: One Adapter. TAda: Two Adapters.}
\vspace{-5pt}
\centering
\begin{threeparttable}
\begin{tabular}{c|c|ccc|ccc}
\toprule
\multicolumn{2}{c|}{\multirow{2}{*}{Method}} & \multicolumn{3}{c|}{SHMOD $\rightarrow$ HZMOD} & \multicolumn{3}{c}{HZMOD $\rightarrow$ SHMOD} \\
\cline{3-8}
\multicolumn{2}{c|}{} & MAE & RMSE & wMAPE & MAE & RMSE & wMAPE \\
\midrule
\multirow{2}{*}{$\text{FF}$} & $\text{OAda}$ & 1.169 & 2.481 & 41.700\% & 0.436 & 1.163 & 59.553\% \\
~ & $\text{TAda}$ & 1.169 & 2.462 & 41.732\% & 0.431 & 1.148 & 58.929\% \\
\hline
\multirow{2}{*}{$\text{OFA}$} & OAda & 1.212 & 2.580 & 43.243\% &0.449 & 1.231 & 61.359\% \\
~ & TAda & 1.195 & 2.532 & 42.632\% & 0.435 & 1.165 & 59.357\% \\
\bottomrule
\end{tabular}
\vspace{-10pt}
\end{threeparttable}
\label{table:cross_city}
\end{table}

\section{Conclusion}
In this paper, we introduce ODMixer, a fine-grained spatial-temporal MLP architecture for metro OD prediction problem. Our ODMixer learns the short-term temporal relations of OD pairs by incorporating the Channel Mixer. The Multi-view Mixer efficiently captures OD pair relations from both origin and destination perspectives. With the integration of BTL, our ODMixer can perceive long-term traffic changes. Experimental results represent ODMixer's outstanding performance on two large-scale datasets. Future directions for ODMixer involve incorporating additional city information, such as urban population distribution, regional composition, and Point of Interest (POI). Moreover, enhancing the model’s ability to learn the general pattern of traffic flow can improve its migration capability and scalability. Deploying ODMixer in actual metro systems can facilitate the management and optimization of metro operations, thereby enhancing transportation efficiency.

\bibliographystyle{IEEEtran}
\bibliography{IEEEabrv,reference}

\vfill
\begin{IEEEbiography}[{\includegraphics[width=1in,height=1.25in,clip,keepaspectratio]{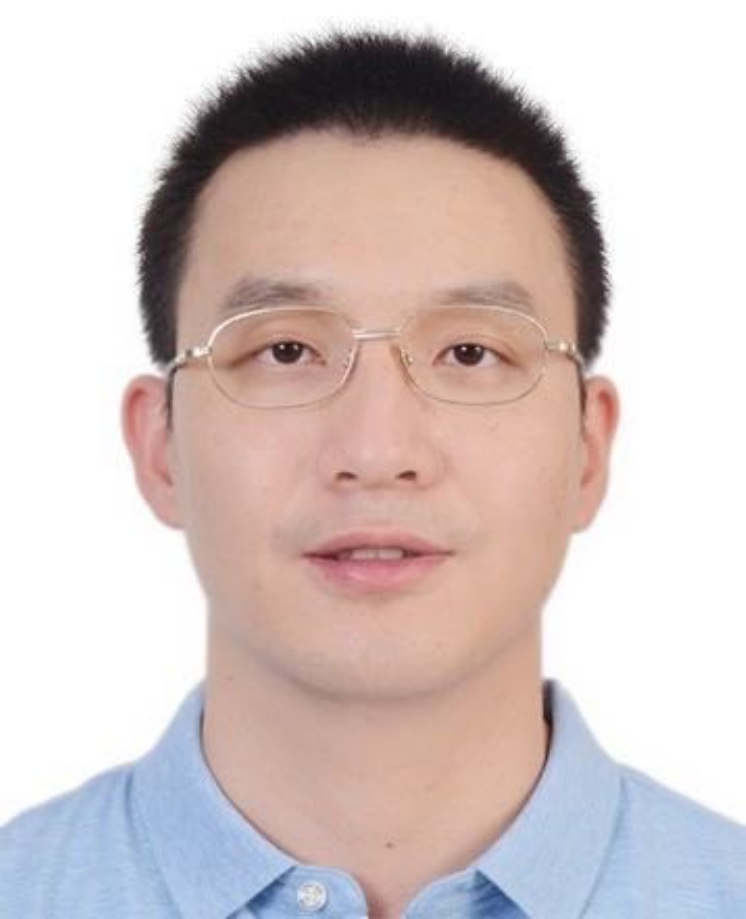}}]{Yang Liu}(Member, IEEE) is currently an associate professor working at the School of Computer Science and Engineering, Sun Yat-sen University. He received his Ph.D. degree from Xidian University in 2019. His current research interests include spatial-temporal representation learning and urban computing. He has published more than 30 papers in top-tier academic journals and conferences such as T-PAMI, T-IP, CVPR, and ICCV, IJCAI, and ACM MM. 
\end{IEEEbiography}

\vfill
\begin{IEEEbiography}[{\includegraphics[width=1in,height=1.25in,clip,keepaspectratio]{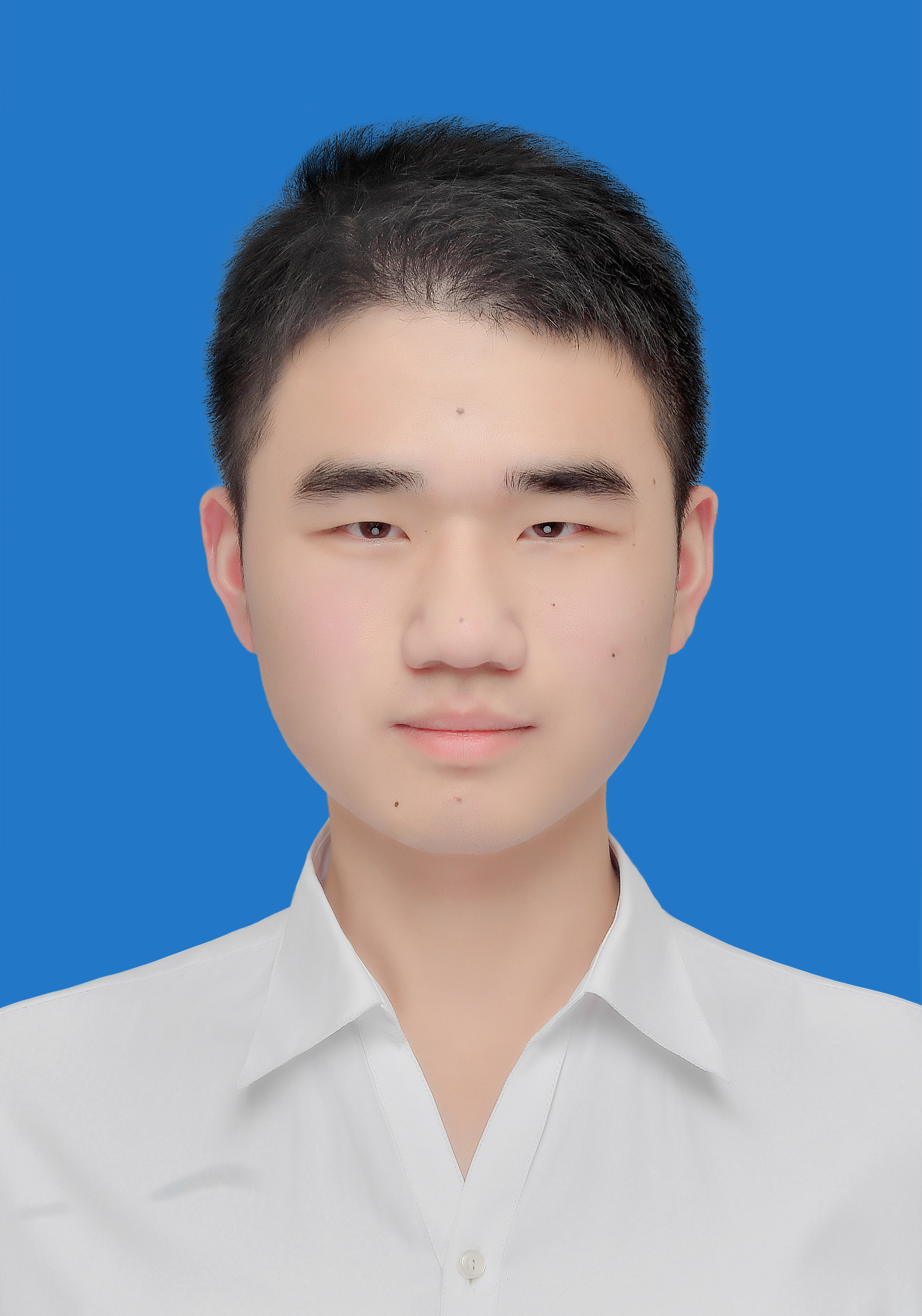}}]{Binglin Chen}
received the B.E. degree from the School of Computer Science and Engineering, Jilin University, Changchun China, in 2022, and he is currently pursuing the Master’s degree in the School of Computer Science and Engineering, Sun Yat-sen University, Guangzhou, China. His current research interests include deep learning, data mining and time series analysis.
\end{IEEEbiography}

\vfill
\begin{IEEEbiography}[{\includegraphics[width=1in,height=1.25in,clip,keepaspectratio]{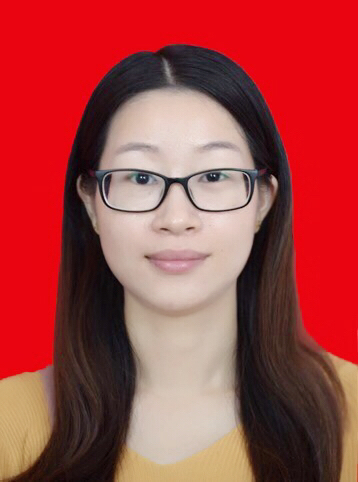}}]{Yongsen Zheng} received her Ph.D. degree in computer science and technology at Sun Yat-sen University, Guangzhou, China, advised by Professor Liang Lin, in 2023. Her current research interests include recommendation systems, knowledge discovery, behavior analysis and machine learning.
\end{IEEEbiography}

\vfill
\begin{IEEEbiography}[{\includegraphics[width=1in,height=1.25in,clip,keepaspectratio]{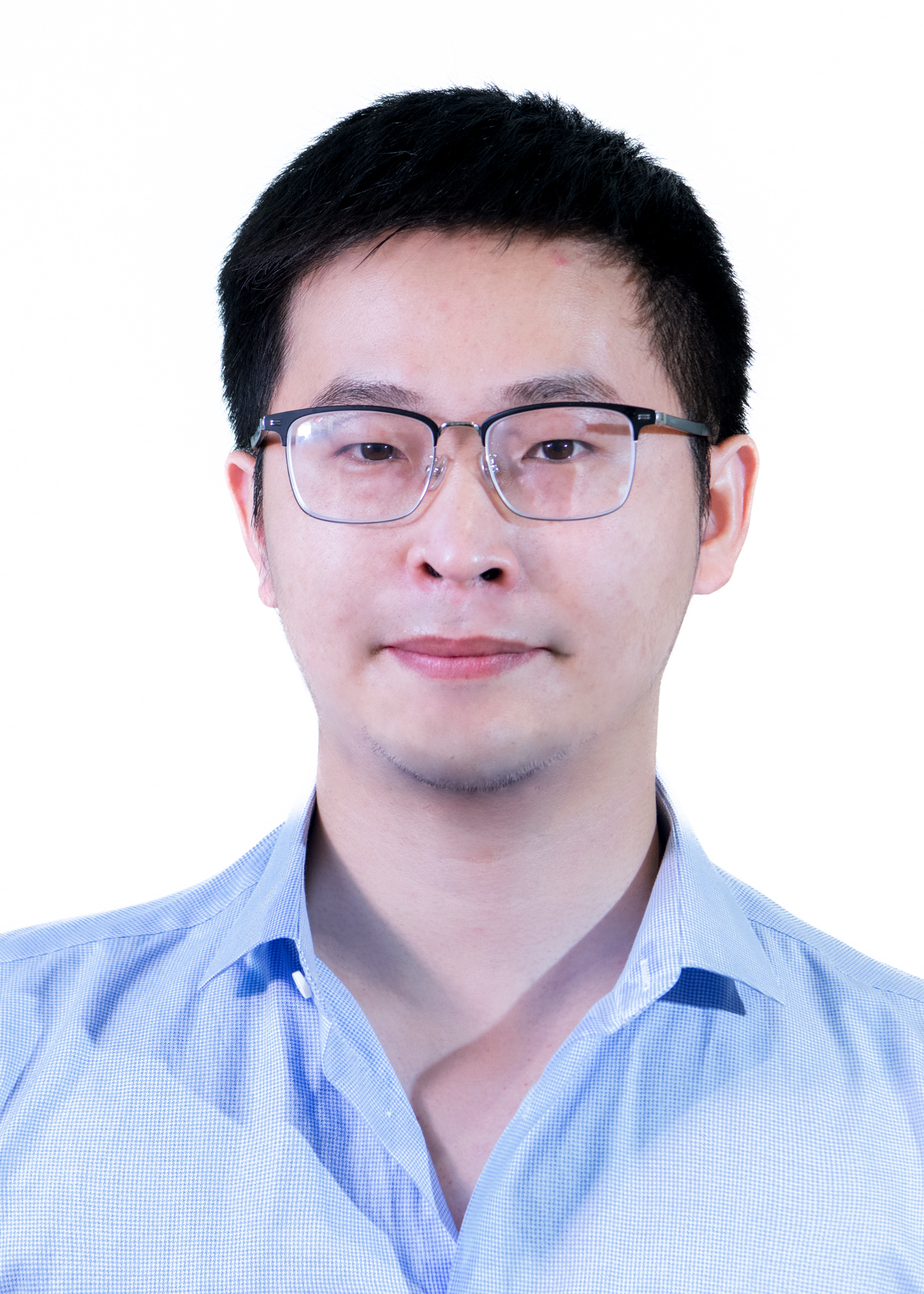}}]{Lechao Cheng} (Member, IEEE) received the Ph.D. degree from the College of Computer Science and Technology, Zhejiang University, Hangzhou, China, in 2019. He is currently an Associate Professor at School of Computer Science and Information Engineering, Hefei University of Technology, Hefei, China. He has contributed more than 50 papers to renowned journals and conferences such as IJCV, TMI, TCyb, TMM, CVPR, ICML, ECCV, AAAI, IJCAI, and ACM MM. His research area centers around vision knowledge transfer in deep learning. 
\end{IEEEbiography}

\vfill
\begin{IEEEbiography}[{\includegraphics[width=1in,height=1.25in,clip,keepaspectratio]{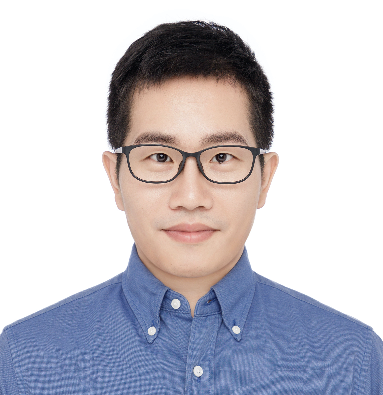}}]{Guanbin Li}(Member, IEEE) is currently an associate professor in School of Computer Science and Engineering, Sun Yat-Sen University. He received his PhD degree from the University of Hong Kong in 2016. His current research interests include computer vision, image processing, and deep learning. He is a recipient of ICCV 2019 Best Paper Nomination Award. He has authorized and co-authorized on more than 70 papers in top-tier academic journals and conferences. He serves as an area chair for the conference of VISAPP. He has been serving as a reviewer for numerous academic journals and conferences such as TPAMI, IJCV, TIP, TMM, TCyb, CVPR, ICCV, ECCV and NeurIPS.
\end{IEEEbiography}

\vfill
\begin{IEEEbiography}[{\includegraphics[width=1in,height=1.25in,clip,keepaspectratio]{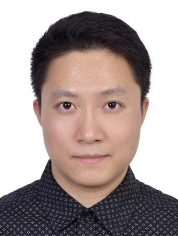}}]{Liang Lin}
(Fellow, IEEE) is a Full Professor of computer science at Sun Yat-sen University. He served as the Executive Director and Distinguished Scientist of SenseTime Group from 2016 to 2018, leading the R\&D teams for cutting-edge technology transferring. He has authored or co-authored more than 300 papers in leading academic journals and conferences, and his papers have been cited by more than 37,000 times. He is an associate editor of IEEE TRANSACTIONS ON
NEURAL NETWORKS AND LEARNING SYSTEMS and IEEE TRANSACTIONS
ON MULTIMEDIA, and served as Area Chairs for numerous conferences such as CVPR, ICCV, SIGKDD and AAAI. He is the recipient of numerous awards and honors including Wu Wen-Jun Artificial Intelligence Award, the First Prize of China Society of Image and Graphics, ICCV Best Paper Nomination in 2019, Annual Best Paper Award by Pattern Recognition (Elsevier) in 2018, Best Paper Dimond Award in IEEE ICME 2017, Google Faculty Award in 2012. His supervised PhD students received ACM China Doctoral Dissertation Award, CCF Best Doctoral Dissertation and CAAI Best Doctoral Dissertation. He is a Fellow of IEEE/IAPR. 
\end{IEEEbiography}
\vfill

\end{document}